\documentclass{article}

\usepackage{PRIMEarxiv}

\usepackage[utf8]{inputenc} 
\usepackage[T1]{fontenc}    
\usepackage{authblk}
\usepackage{hyperref}       
\usepackage{url}            
\usepackage{booktabs}       
\usepackage{amsfonts}       
\usepackage{nicefrac}       
\usepackage{microtype}      
\usepackage{lipsum}         
\usepackage{fancyhdr}       
\usepackage{graphicx}       
\graphicspath{{media/}}     
\usepackage{microtype}
\usepackage{amssymb}
\usepackage{authblk}
\usepackage{multirow}
\usepackage{subcaption}
\usepackage{xcolor}
\usepackage{lineno}
\usepackage{graphicx}
\usepackage[acronym,toc,style=altlist]{glossaries}
\usepackage{setspace} 
\usepackage{pdfpages}
\usepackage{enumitem}
\usepackage{wrapfig}
\usepackage[authoryear]{natbib}

\usepackage{natbib}

\usepackage{longtable}
\usepackage{array}
\usepackage{xcolor}
\usepackage{colortbl}
\definecolor{lightgray}{gray}{0.9}

\hypersetup{
    colorlinks=true,
    linkcolor=blue,
    urlcolor=blue,
    citecolor=blue
}

\title{\texttt{kabr-tools}: Automated framework for multi-species behavioral monitoring
}
\author[1]{Jenna Kline}
\author[2]{Maksim Kholiavchenko}
\author[1]{Samuel Stevens}
\author[3]{Nina van Tiel}
\author[1]{Alison Zhong}
\author[1]{Namrata Banerji}
\author[1]{Alec Sheets}
\author[1]{Sowbaranika Balasubramaniam}
\author[4]{Isla Duporge}
\author[1]{Matthew Thompson}
\author[1]{Elizabeth Campolongo}
\author[5]{Jackson Miliko}
\author[6]{Neil Rosser}
\author[1]{Tanya Berger-Wolf}
\author[2]{Charles V. Stewart}
\author[4,5]{Daniel I. Rubenstein*}
\affil[1]{The Ohio State University} 
\affil[2]{Rensselaer Polytechnic Institute} 
\affil[3]{École Polytechnique Fédérale de Lausanne}
\affil[4]{Princeton University}
\affil[5]{Mpala Research Centre}
\affil[6]{University of Miami}
\affil[*]{corresponding author}

\begin{document}
\maketitle


\begin{abstract}
A comprehensive understanding of animal behavior ecology depends on scalable approaches to quantify and interpret complex, multidimensional behavioral patterns. Traditional field observations are often limited in scope, time-consuming, and labor-intensive, hindering the assessment of behavioral responses across landscapes. To address this, we present \texttt{kabr-tools} (\textbf{K}enyan \textbf{A}nimal \textbf{B}ehavior \textbf{R}ecognition \textbf{Tools}), an open-source package for automated multi-species behavioral monitoring. This framework integrates drone-based video with machine learning systems to extract behavioral, social, and spatial metrics from wildlife footage. Our pipeline leverages object detection, tracking, and behavioral classification systems to generate key metrics, including time budgets, behavioral transitions, social interactions, habitat associations, and group composition dynamics. Compared to ground-based methods, drone-based observations significantly improved behavioral granularity, reducing visibility loss by 15\% and capturing more transitions with higher accuracy and continuity. We validate \texttt{kabr-tools} through three case studies, analyzing 969 behavioral sequences, surpassing the capacity of traditional methods for data capture and annotation. We found that, like Plains zebras, vigilance in Grevy’s zebras decreases with herd size, but, unlike Plains zebras, habitat has a negligible impact. Plains and Grevy's zebras exhibit strong behavioral inertia, with rare transitions to alert behaviors and observed spatial segregation between Grevy’s zebras, Plains zebras, and giraffes in mixed-species herds. By enabling automated behavioral monitoring at scale, \texttt{kabr-tools} offers a powerful tool for ecosystem-wide studies, advancing conservation, biodiversity research, and ecological monitoring.

\end{abstract}



\section{Introduction}

\begin{figure}[!h]
    \centering
    \includegraphics[width=1\linewidth]{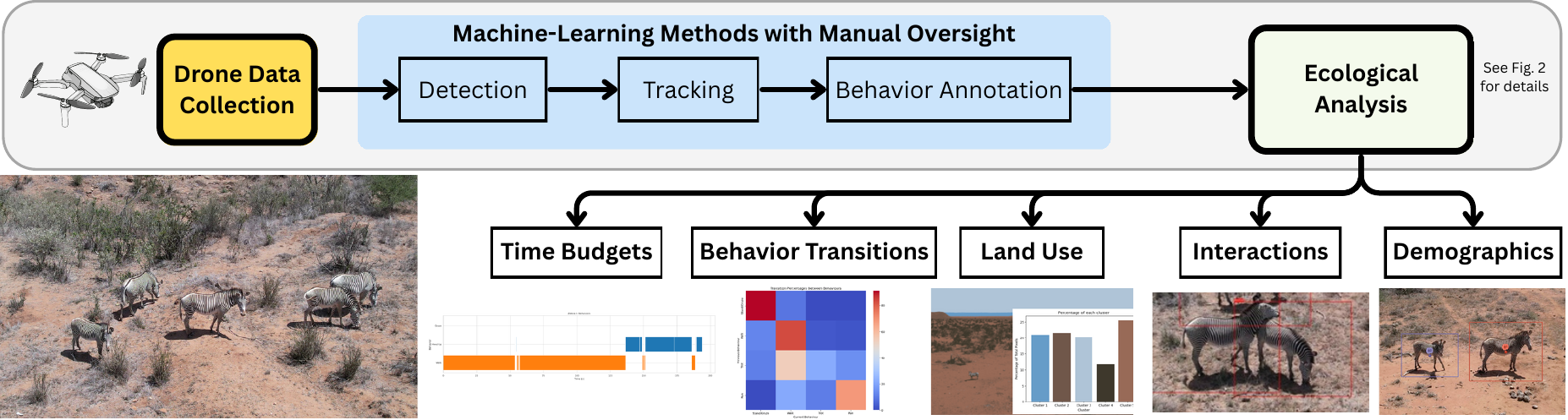}
    \caption{\texttt{kabr-tools} computational framework for automated wildlife behavioral monitoring. The modular pipeline processes drone video through object detection, individual tracking, and machine learning-based behavioral classification to generate ecological metrics including time budgets, behavioral transitions, land and habitat use, social interactions, and demographic data. Framework design enables integration of novel ML models and adaptation across species and study systems. See Fig. \ref{fig:pipeline} for technical details on ML methods with manual oversight.}
    \label{fig:viz_abstract}
\end{figure}

Large-scale wildlife monitoring across diverse ecosystems increasingly requires computational frameworks capable of processing multi-dimensional behavioral datasets to understand behavioral patterns \citep{Koger_Deshpande_Kerby_Graving_Costelloe_Couzin_2023, duporge2025baboonland}. The data-intensive nature of modern ecology, intensified by climate change impacts, habitat fragmentation, and biodiversity loss, demands automated approaches that can scale beyond the temporal and spatial limitations of traditional observation methods \citep{Anderson_Perona_2014, besson2022towards}. While field studies of animal behavior have successfully employed systematic observation methods \citep{Altmann_1974}, these approaches constrain researchers to small-scale, labor-intensive studies that cannot efficiently address landscape-level ecological questions or multi-species comparative analyses.

Traditional behavioral sampling methods, such as focal and scan sampling, have provided the foundation for understanding species ecology and behavioral adaptations through the construction of time budgets \citep{bateson}. However, these approaches face fundamental scalability limitations when applied to contemporary ecological challenges \citep{smith2021observing}. Focal animal sampling, while providing detailed behavioral sequences, restricts observations to single individuals and cannot efficiently capture the simultaneous responses of multiple group members to environmental pressures \citep{Amato_Van_Belle_Wilkinson_2013}. Moreover, by the time the final individual is observed, its state and needs may differ from those observed earlier in the sequence. Scan sampling addresses temporal alignment by recording instantaneous group behavior, but it frequently misses brief but ecologically critical events such as vigilance responses or predator detection behaviors \citep{Amato_Van_Belle_Wilkinson_2013}. Both methods require extensive human observation effort that constrains the spatial scale and temporal duration of studies, limiting our ability to understand behavioral responses to environmental change across communities and landscapes.

Recent advances in remote sensing and computational analysis offer transformative opportunities for scaling behavioral monitoring to meet ecological research needs \citep{hughey2018challenges, Corcoran_Winsen_Sudholz_Hamilton_2021, besson2022towards, Koger_Deshpande_Kerby_Graving_Costelloe_Couzin_2023, kline2025wildwing}. Drone technology provides aerial observation platforms that can simultaneously monitor multiple individuals across groups while minimizing observer effects and visual obstructions \citep{afridi2024noise, pedrazzi2025advancing, schad2023opportunities}. More critically, the integration of computer vision and machine learning enables automated processing of the large video datasets generated by drone surveys, transforming behavioral ecology from observation-limited to data-intensive science \citep{petso2021automatic, zhang2024reliable, saoud2024beyond, axford2024collectively, chan2024yolo, elhorst2025behave, kline2025studying}. These computational approaches can process behavioral information at scales previously impossible with manual methods, enabling more efficient landscape-level monitoring, multi-species comparative studies, and long-term assessment of behavioral responses to environmental change.

Building on this foundation, the Kenyan Animal Behavior Recognition (KABR) project initially developed a dataset of drone-based behavioral observations to train machine learning models for automated behavior classification \citep{kholiavchenko2024kabra}. This foundational work established methodological approaches for drone-based behavioral data collection and demonstrated the feasibility of automated behavior recognition across multiple African savanna species, including Plains zebras (\textit{Equus quagga}), Grevy's zebras (\textit{Equus grevyii}) and reticulated giraffes (\textit{Giraffa reticulata}). However, the original KABR framework was primarily designed for machine learning dataset preparation rather than comprehensive ecological analysis. 
Here, we present an expanded computational framework that extends the KABR project beyond dataset creation to provide comprehensive tools for ecological behavioral analysis, following the application-driven machine learning framework~\citep{rolnick2024application}. Our \texttt{kabr-tools} framework transforms the original machine learning training pipeline into a complete software suite for automated wildlife behavioral monitoring. We build upon the validated methodological foundation established by the original KABR dataset while significantly expanding both analytical capabilities and taxonomic scope. This expanded framework addresses the scalability challenges of traditional methods by providing tools for processing large-scale behavioral datasets across multiple species and ecological contexts.

 We demonstrate the effectiveness of drones in collecting data, paired with machine learning for analysis, by benchmarking the KABR dataset against traditional, ground-based expert observation methods \citep{kline2025kabr-tools-methodology}. We validate our expanded framework using five key metrics and three case studies drawn from the behavioral data in the original KABR video clips and 50 minutes of newly released full-length videos \citep{KABR_Raw_Videos}. We provide worked examples for fourteen new videos, including annotated detections, tracks, behavior labels, and telemetry data \citep{KABR_worked_example}.
Our enhanced framework provides automated extraction of time budgets, behavioral transitions, social interactions, and habitat associations across these diverse study systems. The modular design enables integration of emerging machine learning models and adaptation to new species and study contexts. We demonstrate the ecological applications of this approach through three case studies, which include the computational analysis of anti-predator behavior in Grevy's zebras, transitions in zebra behavior, and inter- and intra-specific social interactions. Our analysis reveals species-specific risk management strategies across habitat gradients. We provide comprehensive documentation \footnote{\href{https://imageomics.github.io/kabr-tools/}{https://imageomics.github.io/kabr-tools/}}, open-source code \footnote{\href{https://github.com/Imageomics/kabr-tools}{https://github.com/Imageomics/kabr-tools}}. All the data is available on HuggingFace, including methodological comparisons\footnote{\href{https://huggingface.co/datasets/imageomics/kabr-methodology}{https://huggingface.co/datasets/imageomics/kabr-methodology}}, and worked examples for three sessions, including full length videos full-length videos\footnote{\href{https://huggingface.co/datasets/imageomics/kabr-full-video}{https://huggingface.co/datasets/imageomics/kabr-full-video}}, detections, tracks, behavior annotations, and associated telemetry\footnote{\href{https://huggingface.co/datasets/imageomics/kabr-worked-examples}{https://huggingface.co/datasets/imageomics/kabr-worked-examples}}. We provide a glossary bridging behavioral ecology and computer vision terminology to facilitate interdisciplinary adoption (Tab. \ref{tab:glossary}). This evolution from dataset creation to a comprehensive computational ecology framework advances automated wildlife monitoring capabilities while providing scalable tools for addressing contemporary conservation biology and ecosystem management challenges. 

\definecolor{lightblue}{RGB}{220,230,241}
\definecolor{lightgreen}{RGB}{220,240,220}

\begin{table}[h!]
\centering
\footnotesize
\begin{tabular}{>{\bfseries}m{3.5cm}m{11.7cm}}
\toprule
\rowcolor{lightblue}
\textbf{Computer Science Term} & \textbf{Definition} \\
\midrule
Annotation & Manual data labeling process – used to train ML models or classify behaviors in datasets \\
\rowcolor{lightblue}
CVAT & Computer Vision Annotation Tool – open-source platform for annotating videos and manually labeling animal behaviors \citep{CVAT} \\
Head Classes & Common behaviors with many training examples (e.g., Walk, Graze, Head Up) \\
\rowcolor{lightblue}
KABR & Kenya Animal Behavior Recognition project – a computational framework for automated animal behavior analysis from drone footage \citep{kholiavchenko2024kabra} \\
Machine Learning (ML) & A type of AI where algorithms learn patterns from data and improve over time without explicit programming. Used in ecology to automate behavior analysis. \\
\rowcolor{lightblue}
Mini-scene & Cropped video sequence focused on a single animal, extracted from drone footage and centered on the individual \\
Tail Classes & Rare behaviors with few training samples (e.g., Fighting), harder to classify \\
\midrule
\rowcolor{lightgreen}
\textbf{Ecological Term} & \textbf{Definition} \\
\midrule
Animal Behaviour Pro & Smartphone app for logging field observations with timestamps and behavior categories \citep{animalbehaviourpro} \\
\rowcolor{lightgreen}
Drone-based Sampling & Behavior is extracted from drone footage, either manually or via AI. Enables focal sampling of all individuals simultaneously. \\
Ethogram & A catalog of species-specific behaviors with standardized descriptions (See example in Tab. \ref{tab:combined_ethogram})\\
\rowcolor{lightgreen}
Focal Sampling & Continuous observation of a single individual recording all behaviors for a predetermined time \citep{Altmann_1974} \\
Ground-based Sampling & Observer records behavior manually (e.g., via binoculars) using focal or scan sampling \citep{Altmann_1974} \\
\rowcolor{lightgreen}
Instantaneous Sampling & Behavior is recorded only at specific observation instants, often using scan sampling \\
Scan Sampling & Recording behavior of all visible group members at regular intervals \citep{Altmann_1974} \\
\rowcolor{lightgreen}
Time Budget & Proportion of time animals spend on various activities like feeding or moving \\
\bottomrule \\
\end{tabular}
\caption{Glossary of Computer Science and Ecological Terms and Tools}
\label{tab:glossary}
\end{table}

\newpage
\section{Methodology}

\begin{figure}
    \centering
    \includegraphics[width=1\linewidth]{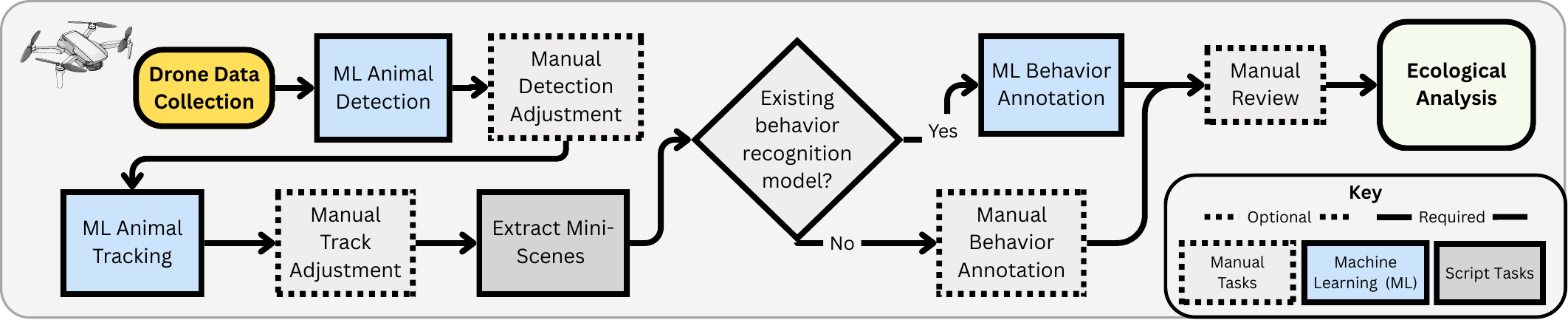}
    \caption{\texttt{kabr-tools} video processing pipeline for drone-based behavioral analysis. Raw drone footage undergoes automated animal detection and tracking, followed by mini-scene extraction to create individual-focused video segments. Manual behavioral annotation using standardized ethograms produces continuous behavioral records, with quality control measures ensuring dataset accuracy and consistency.}
    \label{fig:pipeline}
\end{figure}

We present an end-to-end framework for drone-based behavioral ecology that integrates scalable video processing, expert annotation, machine learning classification, and ecological analysis. Our methodological design prioritizes automation, ecological validity, and minimal disturbance to wildlife. The \texttt{kabr-tools} pipeline, illustrated in Fig. \ref{fig:pipeline}, transforms raw drone footage into behavior-annotated videos. 
We validate our proposed drone-based behavioral analysis pipeline by collecting data on Grevy's zebras, Plains zebras, and reticulated giraffes at Kenya's Mpala Research Centre, using standardized protocols to minimize animal disturbance (Section \ref{data}). We describe the pipeline in detail (Section \ref{videoprocessingpipeline}), including mini-scene extraction (Section \ref{mini-scene}) and the video annotation protocol (Section \ref{video_annotation}). We describe our comprehensive multi-tiered validation strategy (Section \ref{evaluation_method}), which includes machine learning evaluation, cross-method comparisons, and application-driven ecological tasks \citep{rolnick2024application}. We demonstrate that our modular framework enables reproducible, non-invasive behavioral monitoring at scale and provides a template for extending AI-driven analysis to other ecological systems.

\begin{table}[t]
\scriptsize
\centering
\begin{tabular}{@{}llr@{}}
\toprule
\textbf{Category} & \textbf{Metric} & \textbf{Value} \\
\midrule
\multirow{4}{*}{\textbf{Dataset Overview}} 
& Study Period & January 11-21, 2023 \\
& Location & Mpala Research Centre, Laikipia, Kenya \\
& Animals observed & 97+ individuals \\
& Mini-scene count & 969 (807 manually annotated, 162 automatically annotated with ML)\\
\midrule
\multirow{4}{*}{\textbf{Species Distribution}} 
& Plains Zebra & 378 mini-scenes (39.0\%) \\ 
& Grevy's Zebra & 275 mini-scenes (28.4\%) \\ 
& Zebra (species unspecified) & 198 mini-scenes (20.4\%) \\ 
& Reticulated Giraffes & 118 mini-scenes (12.2\%) \\ 
\midrule
\multirow{2}{*}{\textbf{Group Demographics}}  
& Group Size Range & 2-16 individuals \\
& Mean Group Size & 6.1 individuals \\
\midrule
\multirow{4}{*}{\textbf{Drone Sessions}} 
& Total Sessions & 17 \\
& Session Duration Range & 5 - 52 minutes \\
& Typical Session Length & 24 minutes \\
& Recording Time Range & 9:30 AM - 4:30 PM \\
\midrule
\multirow{3}{*}{\textbf{Data Files}} 
& Detections & Bounding boxes locating and tracking animals through the video \\
& Mini-scenes & Cropped video clips created from detections focused on a single individual animal \\
& Behavior & Behavior label associated with each frame in the mini-scene \\
& Drone telemetry & GPS, altitude, heading, and speed  \\
& Drone Session Summary & Session date-time, species, herd size, field notes\\
& Concurrent ground-based scan and focal data & 8 sessions (2 giraffe, 4 Grevy's, 2 Plains)  \\
\bottomrule \\
\end{tabular}%
\caption{Summary of dataset used to validate \texttt{kabr-tools} pipeline and perform case studies analysis. Contains behavior data from eight concurrent scan and focal sampling sessions, 969 annotated mini-scenes obtained from drone footage, and drone session summary.}
\label{tab:dataset_summary}
\end{table}

\subsection{Data Collection for Evaluation}
\label{data}

We collected the data at the Mpala Research Center in Laikipia, Kenya in January 2023, summarized in Tab. \ref{tab:dataset_summary}.
First, we collected ground-based scan and focal sampling to familiarize the field observers with behavior observations and refine the ethograms (see Appendix \ref{ethogram}). 
Next, we manually flew a series of drone missions to collect video footage of animal behaviors, while also conducting simultaneous ground-based focal sampling for selected missions. Acoustic guidelines were used to mitigate disturbance during flight operations \citep{duporge2021determination}.
We collected behavior data of three ungulate species: Grevy's zebras, plains zebras, and reticulated giraffes across diverse habitat types, including open savanna, mixed woodland, and acacia woodland environments.  When possible, demographic information, including age and sex classification, was recorded for the herds. We conducted 19 drone missions using a DJI Air 2S drone, collecting 10.5 hours of video (54 GB) in 4K and 5.4K resolution at 30 frames per second. All research was conducted under Institutional Animal Care and Use Committee approval (Princeton University IACUC 1835F) and Kenya's National Commission for Science, Technology \& Innovation research license (NACOSTI/P/22/18214).

\begin{wrapfigure}{l}{0.45\textwidth}
\vspace{-12pt}
\centering\includegraphics[width=1\linewidth]{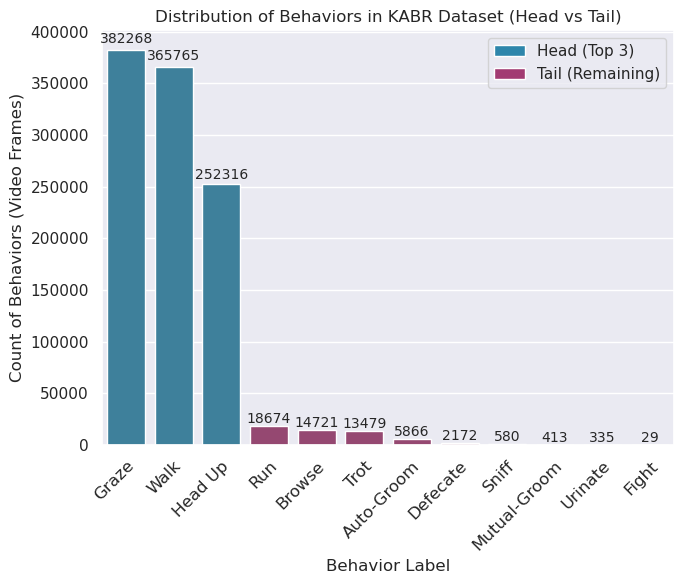}
    \caption{Distribution of behaviors in the dataset. Head classes constitute the majority of the dataset, which include \textit{Graze}, \textit{Walk}, and \textit{Head Up}.}
    \label{fig:kabr-dataset}
\end{wrapfigure}

The dataset includes ground-based sampling data and drone videos capturing ungulate behaviors (Tab. \ref{tab:dataset_summary}). Ground-based scan and focal behavior data was collected for eight sessions to validate the automated classification approaches. We use 969 mini-scenes from 17 drone recording sessions annotated with our pipeline to validate our methodology and perform the case study analysis.
 The dataset contains twelve distinct behavior categories, shown in Fig. \ref{fig:kabr-dataset}). Detailed definitions of the behavioral categories are summarized Appendix \ref{ethogram}. The top seven categories occurring with the greatest frequency (\textit{Graze, Walk, Head Up, Run, Browse, Trot, Auto-Groom}, as well as \textit{Out of Sight}) are used for statistical analysis, including fine-tuning the KABR X3D model \citep{kabr_x3d_model}. 
 Primary behaviors common to all species \textit{Walk} (forward locomotion with at least three consecutive steps), \textit{Head Up} (head raised above shoulder height with ears oriented forward, indicating vigilance or scanning), and \textit{Auto-groom} (self-directed grooming using mouth or hind limb). Secondary behaviors include species-specific activities such as \textit{Browse} for giraffes, describing animals with their heads up eating leaves, and \textit{Graze} for zebras, describing animals with their heads down eating grass. Social interactions, including \textit{Mutual-Groom} and \textit{Fight}, were recorded when observed but occurred too infrequently for statistical analysis. 

\newpage
\subsection{Video Processing Pipeline}
\label{videoprocessingpipeline}

Our video processing methodology transforms raw drone footage into annotated behavioral datasets suitable for machine learning analysis. 
This approach enables simultaneous focal sampling of multiple individuals while maintaining the detailed behavioral records traditionally achieved through single-animal focal sampling. 
The pipeline consists of two main stages. First, the raw videos are processed into mini-scenes (Section \ref{mini-scene}). Next, the mini-scenes are labeled with behaviors (Section \ref{video_annotation}). Quality control measures ensure annotation consistency and accuracy through inter-annotator reliability testing and expert review. Ultimately, the resulting dataset is utilized for ecological analyses.
%


\subsubsection{Mini-scene Extraction}
\label{mini-scene}

\begin{wrapfigure}{r}{0.45\textwidth}
    \centering
\includegraphics[width=\linewidth]{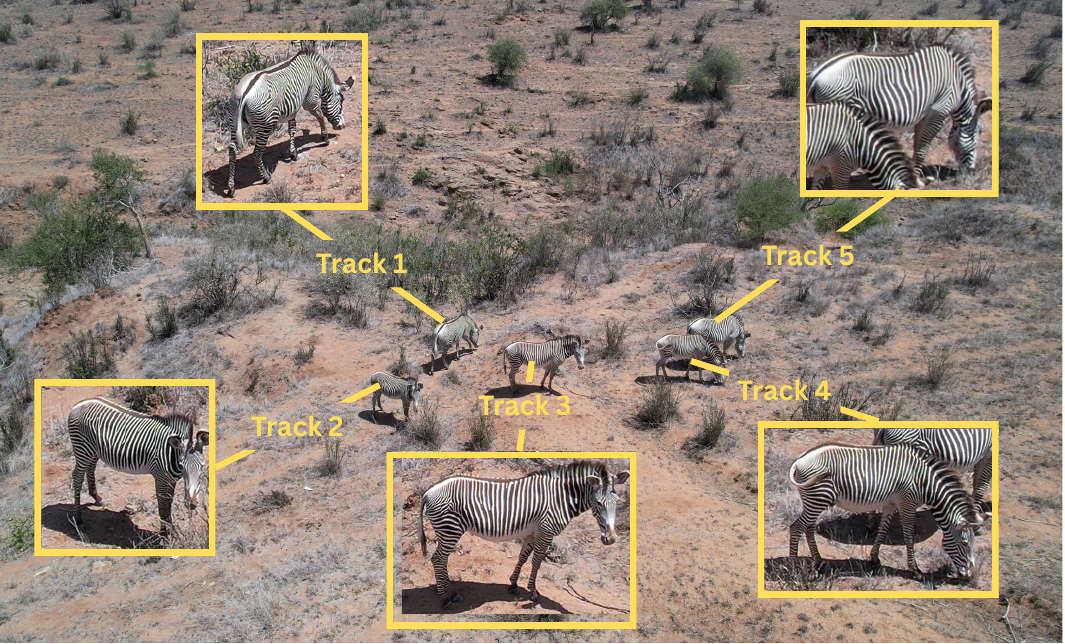}
    \caption{A mini-scene is a sub-image cropped from the drone video footage centered on and surrounding a single animal. Mini-scenes simulate the camera as well-aligned with each individual animal in the frame, compensating for the movement of the drone and ignoring everything in the large field of view but the animals' immediate surroundings. The KABR dataset comprises mini-scenes and their frame-by-frame behavior annotations \citep{kholiavchenko2024kabra}.}
    \label{fig:mini-scene}
\end{wrapfigure}
The initial processing stage begins with object detection to identify all animals in the video frame, followed by continuous tracking to maintain individual identity across frames. From these tracked individuals, we extract \textit{mini-scenes}: cropped video segments that focus on single animals while preserving behavioral context (Fig. \ref{fig:mini-scene}). 
Mini-scenes simulate focal sampling conditions for multiple individuals simultaneously by maintaining each animal at the center of its respective video frame throughout the observation period. Our processing pipeline employs a YOLO-based convolutional neural network for object detection \citep{yolo_2023}, a state-of-the-art model pre-trained on the COCO dataset \citep{coco2015}. 
Detected animals are tracked continuously across frames using spatial proximity and visual feature matching, maintaining consistent individual identification even during temporary occlusions or when animals move close together.
The system dynamically crops the full video frame around each tracked individual while preserving sufficient surrounding context for behavioral classification. This approach compensates for drone movement and camera angle changes, ensuring consistent framing regardless of the animal's position within the larger field of view. 
Quality control measures included manual checks identified evident tracking errors where the algorithm switched between individuals, which were excluded. 
We also filter out mini-scenes consisting of less than 90 frames (i.e. 3 seconds at 30 fps) to ensure a sufficient duration to reliably classify behavior for the focal animal. This duration threshold of three seconds was empirically determined to provide sufficient temporal context for accurate behavioral classification while maximizing data retention. Our manual review and adjustment capabilities corrected tracking errors and refine detection boundaries when necessary. 

\subsubsection{Video Annotation Protocol}
\label{video_annotation}

Once the mini-scenes are extracted, they are annotated with behavior labels frame-by-frame using standardized behavioral labels. This creates continuous behavioral records for each tracked individual. These behavioral labels may be produced manually using an annotation tool such as CVAT, or automatically if a suitable behavior recognition model is available, such as KABR X3D \citep{kabr_x3d_model} or YOLO-Behaviour \citep{chan2024yolo}. To build our dataset, we conducted manual behavior annotations using the Computer Vision Annotation Tool (CVAT). CVAT is an open-source platform enabling precise frame-by-frame labeling of video data \citep{CVAT}. This annotation produces continuous behavioral records analogous to traditional focal sampling, with every frame labeled according to standardized species-specific ethograms (Appendix \ref{ethogram}). 
The annotation protocol requires labeling every frame with the focal individual's current behavioral state, creating temporally precise behavioral sequences. We provide a CVAT user guide developed through this process in our Github repository.

Our annotation team consists of nine trained individuals: four field observers who collected the original ground-based data and five additional assistants who underwent comprehensive training to ensure consistency. Rigorous quality control measures were used to ensure annotation accuracy and consistency. Our standardized training program includes intensive review of species-specific behavioral definitions with video examples, technical instruction on the CVAT interface, and practice annotation sessions until achieving greater than 90\% agreement with expert annotations. We employed weekly calibration sessions throughout the annotation period to address interpretation drift and maintain consistency across all annotators. These included random double-annotation of 10\% of mini-scenes to monitor inter-annotator reliability (achieving $\kappa=0.88$ for primary behavioral categories), weekly calibration sessions to address any annotation drift, and final expert review by field-experienced team members for all completed annotations. This multi-layered approach ensured that the resulting behavioral dataset met the precision standards required for both traditional ethological analysis and machine learning model training.

\subsection{Evaluation Methodology}
\label{evaluation_method}
Our evaluation framework integrates cross-method sampling comparisons and application-driven ecological tasks to assess the effectiveness of our drone-based behavioral analysis pipeline. In Section \ref{drone_eval}, we validate ML-enabled drone-based observations against established ground-based methods by comparing behavior distributions and detection sensitivity across three approaches. (1) scan versus focal sampling (ground-based), (2) focal sampling (ground versus drone), and (3) manual versus ML annotations. In Section \ref{eval_ecologicaltask}, we demonstrate the utility of our pipeline through three case studies, each aligned with key ecological tasks: time budget analysis, behavioral transitions, social interactions, habitat use, and group composition. These case studies illustrate how automated behavioral data extracted from drone footage can be applied to answer ecological questions at individual, group, and community levels. Together, this multi-level evaluation framework validates the technical reliability and ecological relevance of drone-based behavior monitoring systems, enabling scalable, non-invasive observation of wildlife in natural habitats.

\subsubsection{Drone-based Observation Validation}
\label{drone_eval}

We employed multiple analytical approaches to validate ML-enabled drone-based behavior observations against traditional ground-based scan and focal sampling methods. We aim to capture differences in out-of-field annotation protocol with the in-field data collection, since replaying videos typically allows annotators to capture behavioral transitions with greater precision. First, we compared the results from simultaneous ground-based scanning and focal sampling sessions that observed the same herd. Next, we compared the results from simultaneous ground-based focal sampling and drone-based focal sampling, observing the same individual during the same time. Finally, we compared the results of manually annotated drone footage with those of ML-annotated drone footage. To measure the agreement between the methods, we use confusion matrices to quantify the agreement between simultaneous observations, revealing method-specific detection capabilities for different behaviors.

To conduct ground-based observations in the field, observers worked in pairs. One partner observed the behaviors using a pair of binoculars and called out their observations. The other partner recorded the behaviors using the AnimalBehaviorPro app \citep{animalbehaviourpro}. Scan sampling captured data on the whole herd every two minutes, while focal sampling recorded the behavior of a single individual continuously.
Animals were monitored simultaneously using both scan and focal sampling methods (Tab. \ref{tab:scan_focal}). 
We used timestamp data captured with the AnimalBehaviorPro app to match the behavior observations temporally.
To allow for comparison to the continuous focal sampling data, the discrete scan sampling observations were converted to continuous observations propagated across the subsequent two-minute period.

\begin{figure}
    \centering
    \includegraphics[width=1\linewidth]{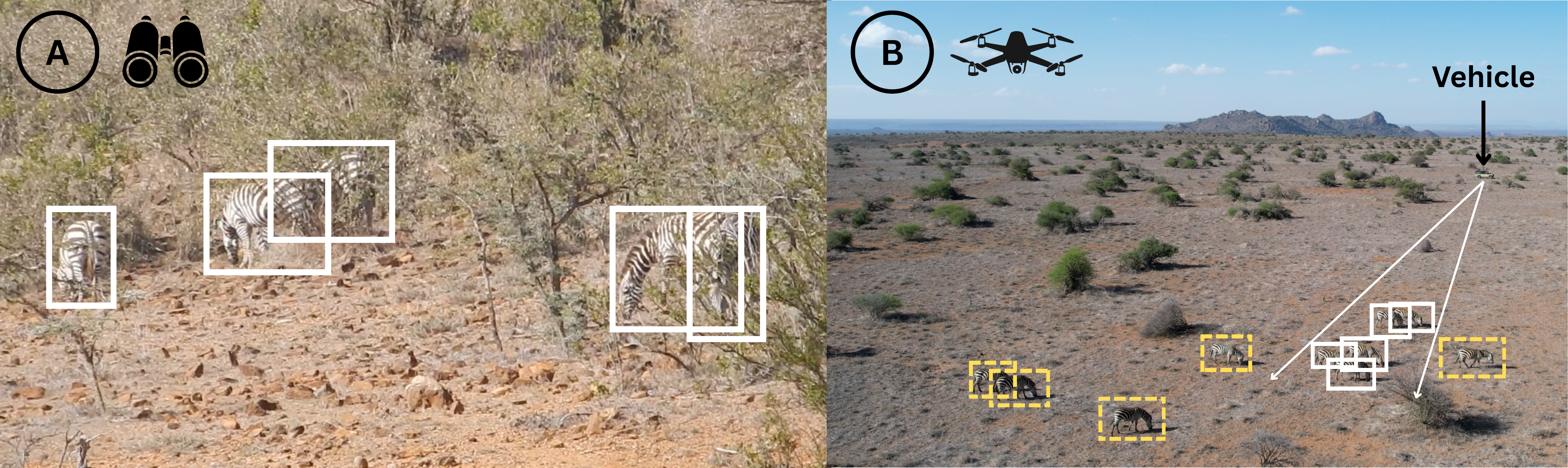}
    \caption{Comparison of ground-based (left) and drone-based (right) perspectives of the same group of plains zebras. While ground-based observations captured only five visible individuals (white solid outline) due to occlusion from vegetation, the drone view clearly revealed all ten (yellow dotted outline), illustrating the advantage of aerial monitoring for behavioral data collection.}
    \label{fig:groundvsdrone}
\end{figure}

During select drone missions, we recorded zebra behaviors using focal sampling, allowing for a comparison of the behavior data gleaned from the drone footage with data observed from the ground (Tab. \ref{table:focal_ind_zebras}). 
We identified and matched the individual zebras between ground observations and drone footage using distinctive physical characteristics, including unique stripe patterns, scars, body size, and sex-specific features through standardized ground-level photography from multiple angles (Fig. \ref{fig:groundvsdrone}). Digital photo catalogs were created for each observation session with GPS coordinates and timestamps to facilitate matching across data collection platforms. Ground-based focal sampling and drone video collection were synchronized using GPS timestamps and detailed field notes recording drone flight initiation and termination times. Video annotations were temporally aligned with ground observations using metadata from both data streams. Only periods with complete temporal overlap between ground focal sampling and corresponding drone mini-scenes were included in comparative analyses. To ensure valid comparisons between drone and ground-based behavioral datasets, we applied four harmonization steps (Tab. \ref{tab:data-harmonization}). We first filtered out periods where individuals were not visible in either method. Then, we harmonized behavior categories, validated individual identity matches, and matched temporal resolutions by aggregating drone data. These steps minimized bias and improved alignment between datasets for robust comparative analysis.

In our previous work, we evaluated four ML models to automatically classify behavior from drone videos: I3D \citep{carreira2017quo}, SlowFast \citep{slowfast_2019}, X3D \citep{feichtenhofer2020x3d}, and UniformerV2 \citep{li2022uniformerv2} \citep{kabrdeepdive}. The X3D model consistently performed the best, so it was used here to evaluate the performance of automatic behavior recognition compared to manually annotated mini-scenes. The X3D model achieves a  91.32\% F-1 score for the three most common behaviors, which include \textit{Graze}, \textit{Walk}, \textit{Head Up}, but significantly lower performance on long-tailed tail classes, consisting of rarer behaviors (50.31\% F-1). We used the X3D model to automatically annotate the behaviors used for Case Study 3 (Section \ref{cs1}).

\begin{table}[t]
\footnotesize
\centering
\begin{tabular}{p{5cm}p{11cm}}
\toprule
\textbf{Data Harmonization Step} & \textbf{Description} \\
\midrule
Visibility Filtering & Excluded time periods when individuals were out of sight (ground focal) or out of frame/occluded (drone focal) to ensure valid time budget comparisons. \\
Behavioral Category Harmonization & Applied consistent behavioral definitions across both ground and drone datasets to allow for direct comparison. \\
Individual Matching Validation & Limited comparative analyses to individuals with clear, unambiguous identification across both methods. \\
Temporal Resolution Matching & Aggregated drone annotations to match the coarser temporal resolution of ground observations when required. \\
\bottomrule \\
\end{tabular}
\caption{Data harmonization steps used to align drone and ground-based behavioral datasets.}
\label{tab:data-harmonization}
\end{table}

\subsubsection{Ecological Tasks and Case Studies}
\label{eval_ecologicaltask}
We focus on five broad ecological tasks: time budget analysis, behavioral transitions, social interactions, land use and movement patterns, and social group composition analysis \citep{hughey2018challenges, smith2021observing, pedrazzi2025advancing}, summarized in Tab. \ref{tab:analysis_summary}. These tasks provide a multifaceted understanding of animal behavior by capturing not just how much time animals spend in different activities, but also how behaviors transition over time, vary across habitats, and are shaped by social context. Time budgets and transition matrices offer standardized metrics to quantify behavioral patterns and flexibility, while drone-enabled detection of interactions and movement pathways reveals dynamics often missed by ground observers. Together, these methods help ecologists link behavior to environmental and social drivers, which is essential for interpreting animal responses to habitat change, human disturbance, and conservation interventions.

\begin{table}[h!]
\centering
\footnotesize
\begin{tabular}{>{\bfseries}m{3.3cm}m{6cm}m{6.2cm}}
\toprule
\textbf{Ecological Task} & \textbf{Description} & \textbf{Key Metrics / Comparisons} \\
\midrule

Time Budget Analysis & 
Calculates proportion of time spent in each behavioral state. Excludes periods when the animal was not visible. &
\begin{itemize}[leftmargin=*,noitemsep]
    \item Time Budget$_i = \frac{\sum t_i}{\sum t_{\text{visible}}} \times 100\%$
    \item Comparison across drone vs. ground observations
\end{itemize} \\

\rowcolor{gray!10}
Behavioral Transitions & 
Examines sequential behavior using transition matrices. &
\begin{itemize}[leftmargin=*,noitemsep]
    \item $P(j|i) = \frac{n_{ij}}{\sum_k n_{ik}}$
    \item Drone vs. ground transition patterns
\end{itemize} \\

Social Interactions & 
Detects social interactions based on spatial overlap and behavioral synchrony in drone footage. &
\begin{itemize}[leftmargin=*,noitemsep]
    \item $>50\%$ bounding box overlap for $>3$ frames
    \item (Optional) Behavioral categorized interactions (e.g., mutual-grooming)
\end{itemize} \\

\rowcolor{gray!10}
Land Use and Movement Patterns & 
Uses GPS and visual classification to link behavior to habitat type. &
\begin{itemize}[leftmargin=*,noitemsep]
    \item Habitat: open grassland, dense woodlands, mixed
    \item Movement paths in annotated landscapes
\end{itemize} \\

Social Group Composition Analysis & 
Analyzes group size, demographics, and spatial structure based on high-resolution video. &
\begin{itemize}[leftmargin=*,noitemsep]
    \item Age-sex categories (adult male/female, subadult, juvenile, infant)
    \item Group size and spatial configuration
\end{itemize} \\

\bottomrule \\
\end{tabular}
\\
\caption{Summary of behavioral and ecological analyses enabled by drone footage}
\label{tab:analysis_summary}
\end{table}

To demonstrate the analytical capabilities enabled by drone-based observation, we conducted three case studies leveraging the tasks (Tab. \ref{tab:case_studies}). Herbivores must balance competing needs to feed and avoid predation. These competing needs cause animals to avoid areas where food is abundant but predation risk is high, known as the landscape of fear phenomenon \citep{riginos2015climate}. We examine this phenomenon in our first case-study, \textit{Grevy’s Zebra Landscape of Fear}, which combines time budgets, habitat use, and social group composition to explore how landscape features and group structure influence perceived risk and anti-predator behavior. We model the likelihood of transitions between locomotor states in the \textit{Zebra State Shift} case study to assess behavioral flexibility and alertness. Finally, the \textit{Cross-Species Encounters} case study uses spatial proximity and behavioral synchrony to detect and classify social interactions, providing insight into mixed-herd dynamics that are difficult to observe from the ground.

\begin{table}[t]
\centering
\footnotesize
\begin{tabular}{>{\bfseries}m{3.5cm}m{7cm}m{4.5cm}}
\toprule
\textbf{Case Study} & \textbf{Description} & \textbf{Ecological Tasks} \\
\midrule

Grevy's Zebra Landscape of Fear & 
Analyzed 368 mini-scenes of Grevy’s zebra groups (2–8 individuals) across open grassland and bush habitats to assess anti-predator behavior. Behavioral labels were generated using a fine-tuned X3D model \citep{kabr_x3d_model}. & 
\begin{itemize}[leftmargin=*,noitemsep]
    \item Time budgets
    \item Land use 
    \item Social group composition 
\end{itemize} \\

\rowcolor{gray!10}
Zebra State Shifts & 
Modeled behavior sequences from 748 mini-scenes to estimate transition probabilities between grazing, walking, trotting, and running. & 
\begin{itemize}[leftmargin=*,noitemsep]
    \item Behavioral transitions 
    \item Time budgets
\end{itemize} \\

Cross-Species Encounters & 
Detected 5,035 inter- and inter-species interactions among 16 individuals using the bounding box overlap method across 157 annotated mini-scenes. & 
\begin{itemize}[leftmargin=*,noitemsep]
    \item Social interaction detections 
    \item Social group composition
\end{itemize} \\

\bottomrule \\
\end{tabular}
\caption{Case studies illustrating use of analytical tasks for drone-based behavioral ecology}
\label{tab:case_studies}
\end{table}
\section{Results }

We evaluate the performance of our end-to-end system for automated wildlife behavior analysis through a combination of machine learning benchmarking, method comparison, and ecological case studies.
In Section \ref{method_comparison}, we report the results of drone-based observations compared to traditional ground-based focal and scan sampling methods, and evaluate the performance of deep learning models for automatic behavior labeling. In Section \ref{casestudy}, we report the findings of our three case studies. The results demonstrate the potential of drone-based, ML-enabled behavioral analysis to match and surpass traditional methods in accuracy, scale, and ecological insight.

\subsection{Comparison of Behavior Data Collection Methods}
\begin{figure}[htbp]
    \centering
    \begin{subfigure}[b]{0.32\textwidth}
        \includegraphics[width=\textwidth]{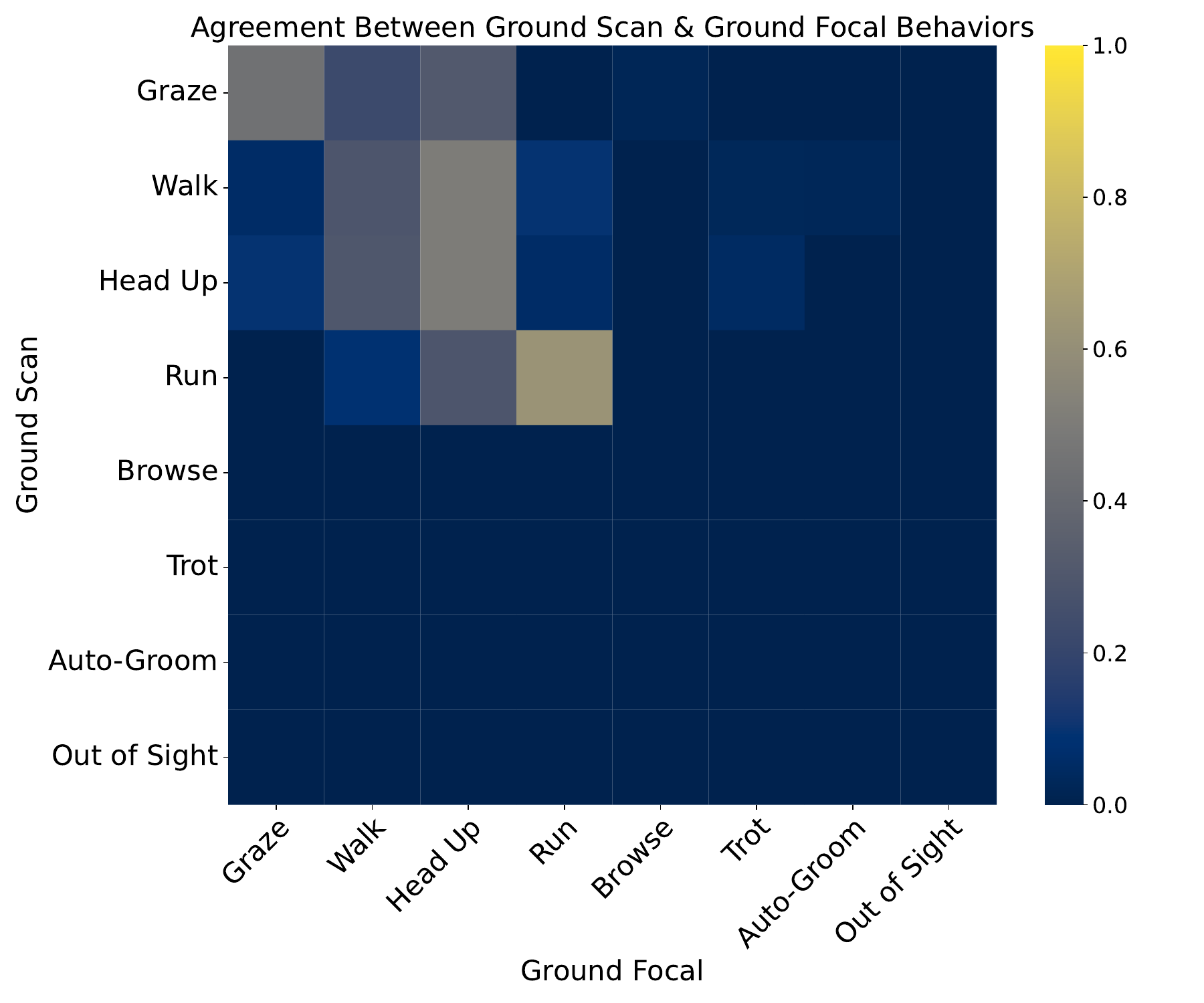}
        \caption{Data yielded from ground-based scan sampling (y-axis) versus the ground-based focal sampling (x-axis) (See Tab. \ref{tab:scan_focal}) for demographic details.}
        \label{fig:scan_focal}
    \end{subfigure}
    \hfill
    \begin{subfigure}[b]{0.32\textwidth}
        \includegraphics[width=\textwidth]{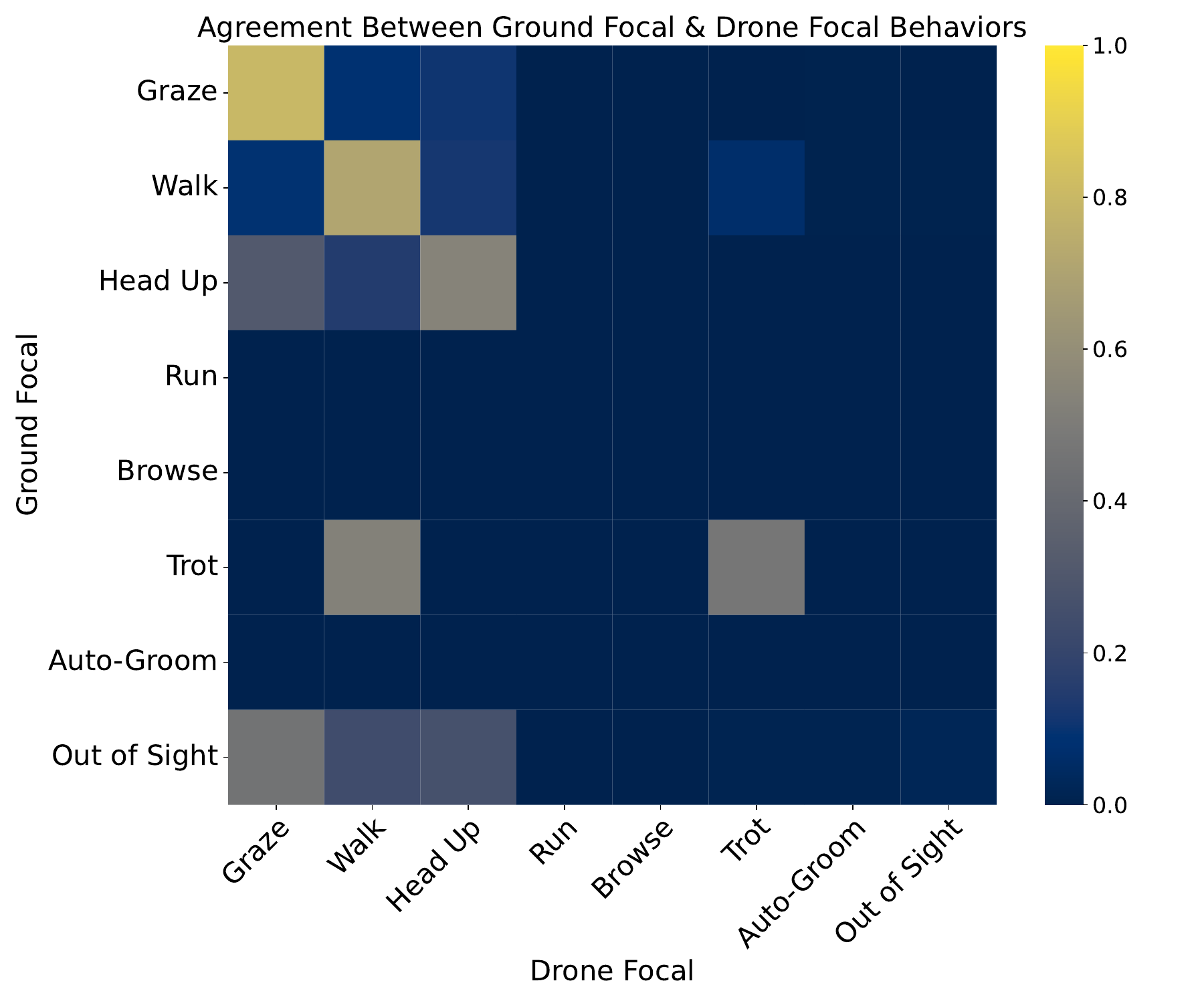}
        \caption{Data yielded from ground-based focal sampling (y-axis) versus drone-based focal sampling (x-axis). (See  Tab. \ref{table:focal_ind_zebras})for demographic details.}
        \label{fig:drone_ground}
    \end{subfigure}
    \hfill
    \begin{subfigure}[b]{0.32\textwidth}
        \includegraphics[width=\textwidth]{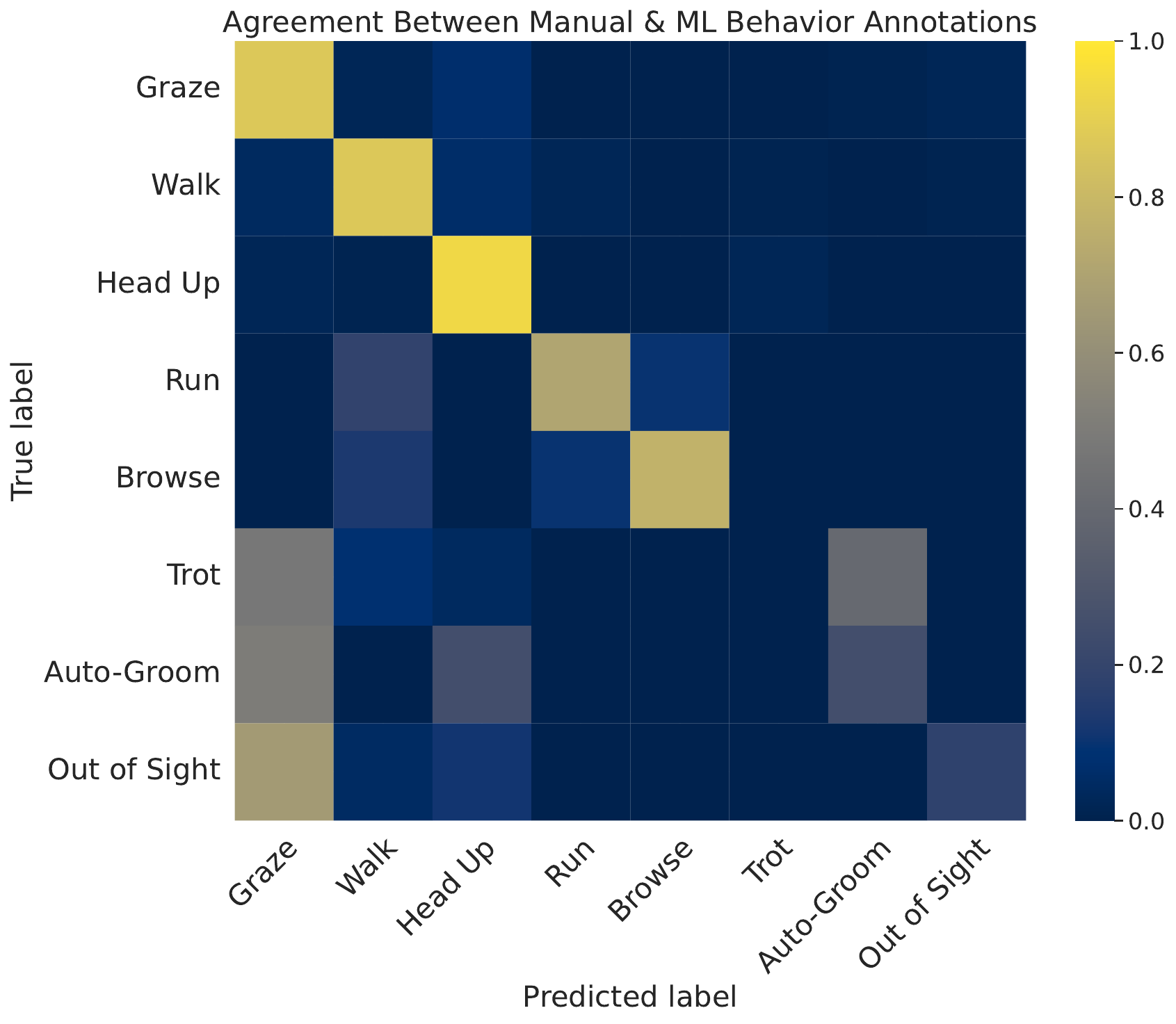}
        \caption{Comparison of manually-labeled behavior data (y-axis) versus automatically labeled behavior data using ML (x-axis).  }
        \label{fig:manual_automatic}
    \end{subfigure}
    \caption{Comparison of behavior data collected using different methodologies applied to the same individual animal during the same time. Drone-based methods enable concurrent focal sampling of all individuals in a herd, and ML models can automatically annotate behaviors with high accuracy--particularly for common behaviors--closely matching manual and ground-based observations while capturing events often missed from the ground. Behavior categories are arranged in descending order by frequency of occurrence. (a) Ground-based scan and focal sampling show stronger agreement for common behaviors, but scan sampling fails to capture rare behaviors. (b) Ground-based focal sampling matches drone-based focal sampling, but misses behaviors when the animals are out of sight from the ground-based observers. (c) Automatic ML behavior annotation shows strong performance compared to manual annotations, especially for common behaviors. }
    \label{fig:cms}
\end{figure}
\label{method_comparison}

To evaluate the effectiveness of drone-based behavioral sampling, we perform three key comparisons: (1) ground-based scan versus ground-based focal sampling (Section \ref{ground_methods}), (2) ground-based focal versus drone-based focal sampling (Section \ref{ground_drone_focal}), and (3) manual versus automated behavior annotations of drone-based observations (Section \ref{ml_manual}) \citep{kline2025kabr-tools-methodology}. We use confusion matrices to compare concordance between these methodologies (Fig. \ref{fig:cms}). Confusion matrices are tabular summaries that quantify agreement between two classification methods by reporting the frequency with which each behavior label from one method matches or differs from the corresponding label in the other. Our comparison confirmed that scan sampling captures concurrent behavior of the entire herd, but captures fewer behavior categories overall compared to individual focal sampling. Drone-based focal sampling reduced out-of-sight time by 14.7\% and captured more behavioral transitions compared to ground-based focal sampling. A comparison of manual versus automated drone annotations showed high agreement for dominant behaviors, confirming the utility of automated pipelines while revealing systematic errors in rare or subtle behaviors. Together, these comparisons demonstrate that drone-based sampling with machine learning provides a powerful complement to traditional methods, combining the comprehensive coverage of scan sampling with the fine granularity of focal sampling, while enabling scalable, automated behavioral analysis at high resolution.

%

\subsubsection{Comparison of ground-based focal and scan sampling}
\label{ground_methods}
Simultaneous scan and focal sampling in the field reveal key differences in behavioral granularity, illustrated in Fig.~\ref{fig:scan_focal}.
Focal sampling captured significantly more behavioral categories than scan sampling (focal: mean $= 6.2 \pm 1.4$ categories; scan: mean $= 3.1 \pm 0.8$ categories; paired t-test: $t_{5} = 4.73$, $p = 0.005$).  Focal sampling captured rare and transient behaviors, including \textit{Trot}, \textit{Auto-Groom}, and \textit{Fight}, which were often missed by the coarser scan sampling approach. Scan sampling captures \textit{Graze}, \textit{Walk}, \textit{Head Up} and \textit{Browse}. When scan sampling captures \textit{Graze}, focal sampling revealed this included \textit{Graze} (44\%), \textit{Walk} (25\%), \textit{Head Up} (28\%), and \textit{Auto-Groom} (2\%).

\subsubsection{Comparison of ground-based focal and drone focal sampling}
\label{ground_drone_focal}
We compare ground-based focal sampling with drone-based focal sampling using temporally aligned video segments from four uniquely identifiable zebras (Tab.~\ref{table:focal_ind_zebras}). 
Temporal overlap for each sampling session ranged from 193.9 to 405.8 seconds for a total of 22 minutes. Drone-based sampling was found to offer substantial advantages in visibility and coverage: ground observers lost sight of focal animals for an average of 23.4\% of the observation time, whereas drones experienced only 8.7\% of occlusion. Ground-based and drone-based focal sampling were largely in agreement for common behaviors, \textit{Graze, Walk, Head Up} (Fig. \ref{fig:drone_ground}). Behavior instances labeled \textit{Trot} in ground-based sampling was frequently labeled as \textit{Walk} in the drone focal sampling.
Ground sampling suffered from extended \textit{Out of Sight} periods, while drone sampling experienced shorter \textit{Out of Sight} periods. When the animals were out of sight of the ground observers, the drone captured \textit{Graze } (46\%), \textit{Walk} (24\%), and \textit{Head Up} (26\%) behaviors. When both methods successfully capture behavior, agreement was highest for common behaviors: \textit{Graze }(80\% concordance), \textit{Walk} (71\% concordance), but lower for brief behaviors: \textit{Head Up} (54\% concordance), \textit{Trot/Run} (47\% concordance). 

\begin{figure}[htbp]
    \centering
    
    \begin{subfigure}[b]{0.9\textwidth}
        \centering
        \includegraphics[width=1\linewidth]{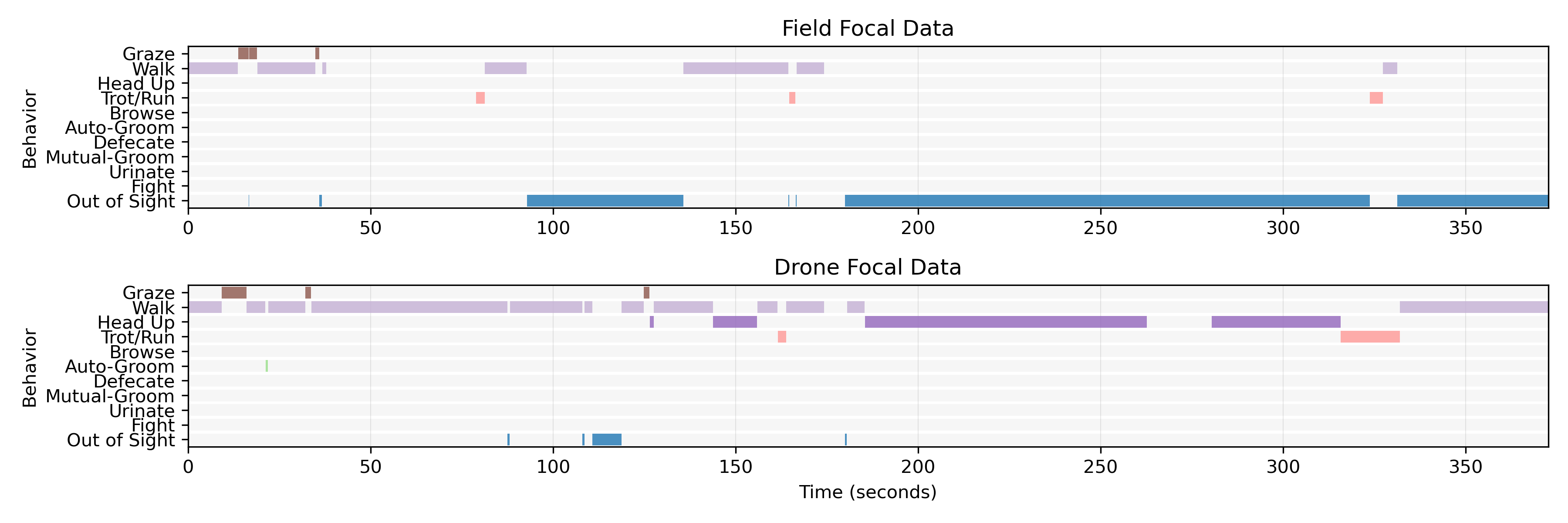}
        \caption{Zebra A Gantt Chart}
        \label{fig:zebraA}
    \end{subfigure}
    \begin{subfigure}[b]{0.9\textwidth}
        \centering
        \includegraphics[width=1\linewidth]{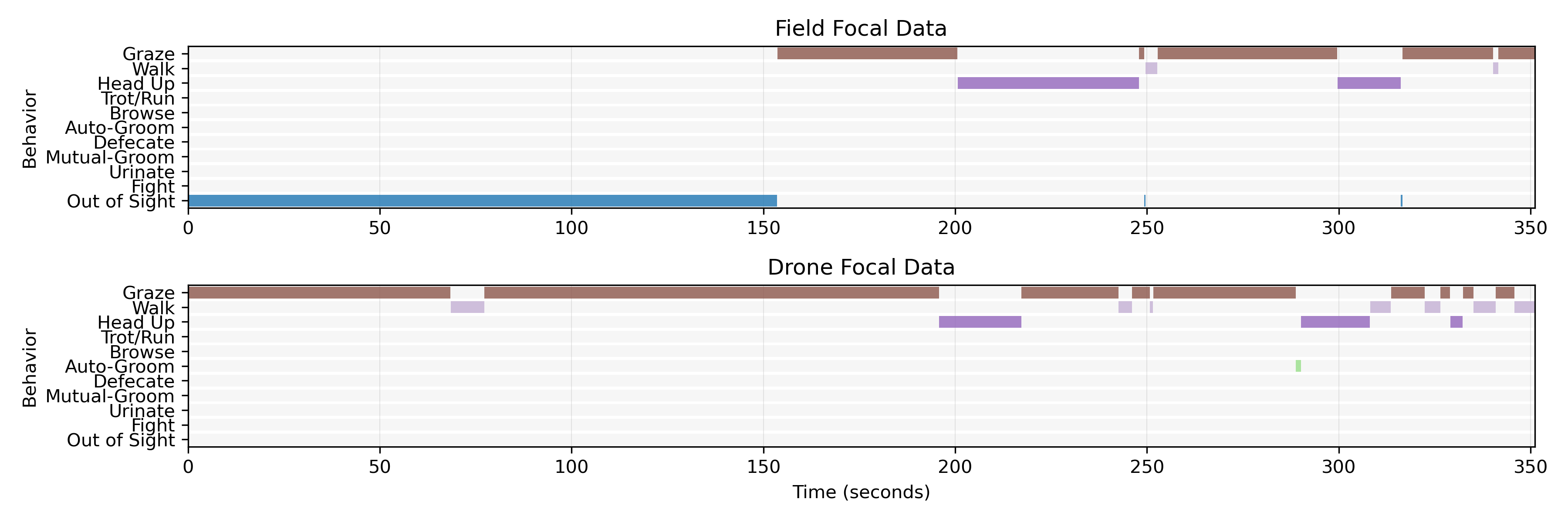}
        \caption{\footnotesize{Zebra B Gantt Chart}}
        \label{fig:zebraB}
    \end{subfigure}

    \begin{subfigure}[b]{0.9\textwidth}
        \centering
        \includegraphics[width=\textwidth]{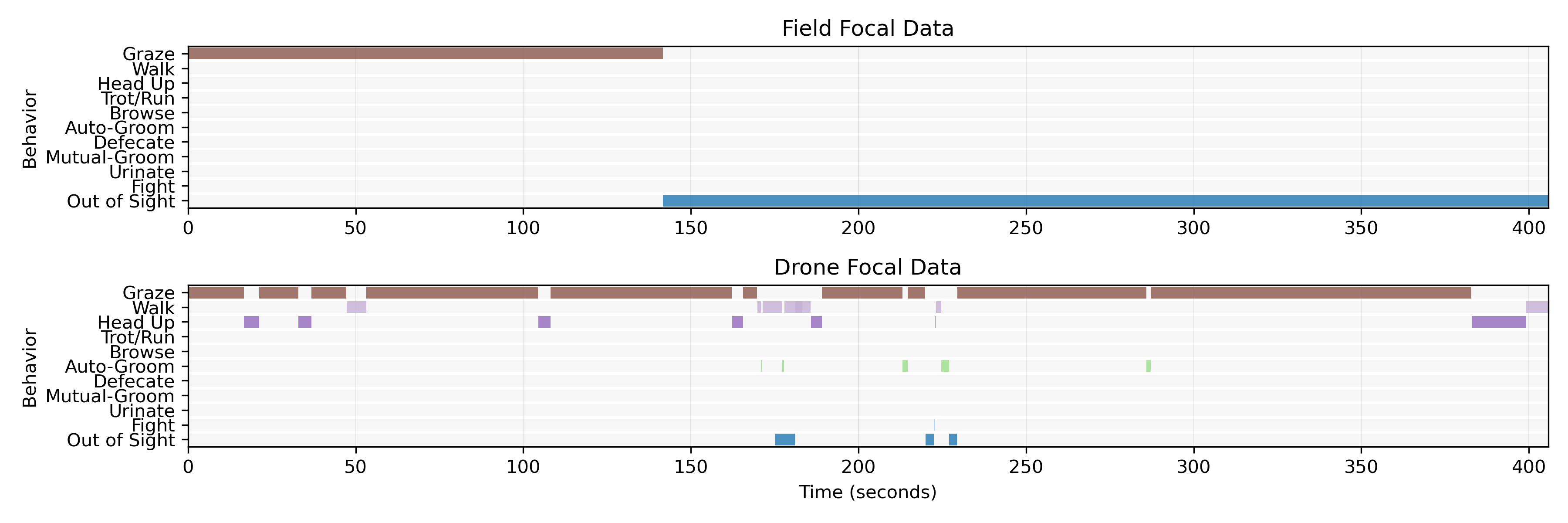}
        \caption{Zebra C Gantt Chart}
        \label{fig:zebraC}
    \end{subfigure}

    \begin{subfigure}[b]{0.9\textwidth}
        \centering
        \includegraphics[width=\textwidth]{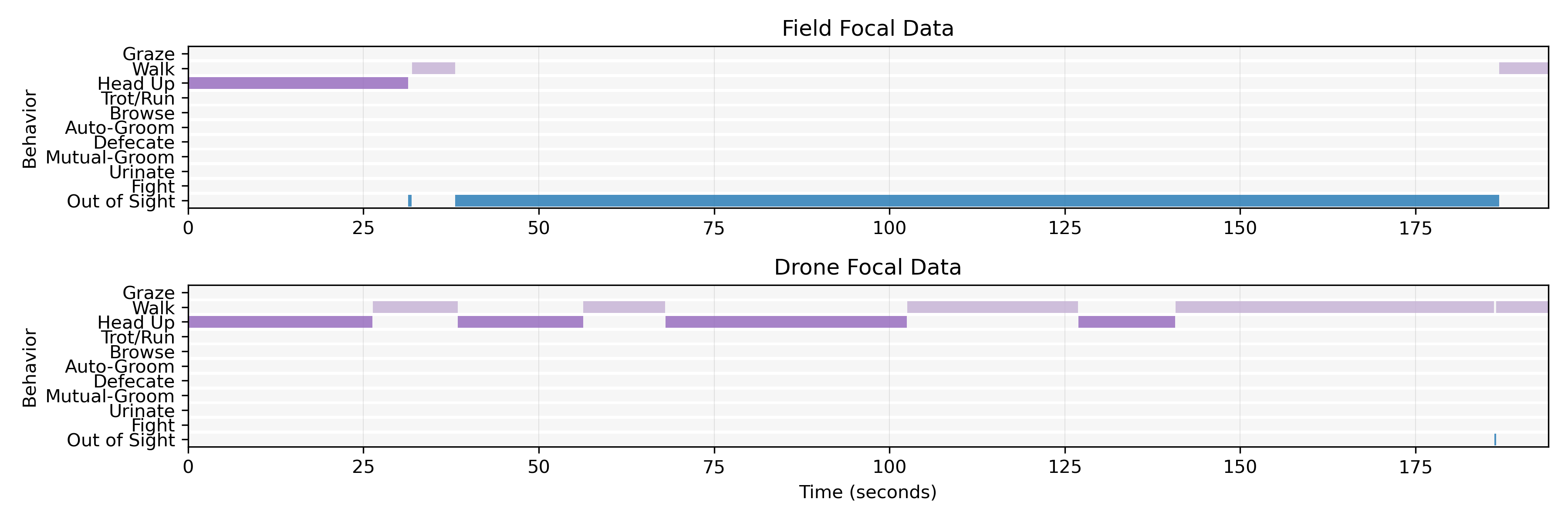}
        \caption{Zebra D Gantt Chart}
        \label{fig:zebraD}
    \end{subfigure}
    \caption{Temporal comparison of behavior captured from ground-based focal sampling versus drone-based focal sampling of four zebras (Tab. \ref{table:focal_ind_zebras}). Drone observations yielded finer granularity and have far fewer instances of \textit{Out of Sight} compared to ground-based observations. Behavior categories are organized in descending order of occurrence.}
    \label{fig:ganttcharts}
\end{figure}

Gantt chart comparisons further demonstrated that drones captured more behavioral transitions and continuous observation windows, providing richer behavioral sequences, as shown in the Gantt charts in Fig. \ref{fig:ganttcharts}. Drone footage consistently captures more behavior transitions than ground observations. For Zebra A, the drone-based focal sampling largely matches the ground-based focal sampling for the first 100 seconds, until the zebra is out of sight of the ground observers. For Zebra B, the individual was out of sight of the ground-based observers for the first 150 seconds, while the drone captured instances of \textit{Graze} and \textit{Walk}. Once Zebra B was in view of the ground observers, the behaviors captured were the same as those of the drone-based observers. Notably, the drone-based observations captured more behavioral transitions and a brief instance of \textit{Auto-Groom}. Similarly, for Zebra C, the drone-based focal observations captured instances of \textit{Graze}, \textit{Walk}, and \textit{Head Up}, with rapid behavioral changes, while ground-based observations only captured \textit{Graze} during the same time. After 150 seconds, Zebra C was out of sight of the ground-based observers, while the animal stayed in view of the drone for 400 seconds, with a few brief instances of occlusion. For Zebra D, the drone-based focal sampling agreed with the ground-based focal observations, although the drone captured the behavior transition from \textit{Head Up} to \textit{Walk} sooner than the ground observers. After 30 seconds, Zebra D was out of sight of the ground-based observers, while the animal stayed in view of the drone for 200 seconds. Note, this small sample size ($n=4$) precludes statistical inference and serves only for qualitative method comparison. Future studies require larger sample sizes for robust statistical analysis.

\subsubsection{Comparison of manually-annotated and automatically-annotated behaviors}
\label{ml_manual}

We assess the performance of automated behavior recognition using the best-performing X3D model against expert manually annotated drone footage. The X3D model is evaluated using 25\% of the mini-scenes for testing, while the remaining 75\% were used for training \citep{kholiavchenko2024kabra}. The confusion matrix (Fig.~\ref{fig:manual_automatic}) compares manual behavior annotations (rows) to predictions generated by the X3D model (columns) across the eight behavioral categories. The X3D model demonstrates high accuracy for core locomotor and foraging behaviors, correctly classifying \textit{Graze}, \textit{Walk}, and \textit{Head Up} with accuracies of 87\%, 87\%, and 94\%, respectively. Moderate performance is observed for \textit{Trot} (71\%) and \textit{Run} (77\%), although some misclassification occurs between these and \textit{Walk}, likely due to overlapping kinematic features.  Less frequent or more visually subtle behaviors such as \textit{Browse} and \textit{Auto-Groom} exhibit higher rates of confusion, particularly with grazing and scanning postures. Misclassifications were most common between visually similar categories and in occluded frames, which were often predicted as \textit{Graze}. For example, \textit{Browse} is frequently predicted as \textit{Graze} (48\%) and \textit{Auto-Groom} is often misclassified as \textit{Graze} (50\%) or \textit{Head Up} (25\%). The \textit{Out of Sight} category is partially misclassified as \textit{Graze} (66\%), suggesting that occluded frames may share visual characteristics with feeding behaviors.

\subsection{Ecological Case Studies}
\label{casestudy}
We demonstrate the capabilities of our behavioral analysis pipeline through three ecological case studies. For Case Study 1, \textit{Grevy’s Zebra Landscape of Fear} (Section \ref{cs1}), we found that vigilance declined with increasing herd size, consistent with the group-vigilance trade-off, but habitat type did not explain variation in behavior, unlike Plains zebras. For Case Study 2, \textit{Zebra State Shifts} (Section \ref{cs2}), zebras showed strong behavioral inertia, with infrequent but detectable transitions to alert behaviors (e.g., \textit{Walk} to \textit{Run}). These patterns offer a foundation for predictive, behavior-adaptive remote sensing systems. For Case Study 3, \textit{Cross-Species Encounters} (Section \ref{cs3}), we analyzed the spatial overlap among Grevy's zebras, Plains zebras, and reticulated giraffes in a mixed-species herd. Intra-species proximity was highest for zebras, while interactions between zebras and giraffes were not detected, indicating spatial segregation between species despite sharing a grazing area.

\subsubsection{Case Study 1: Grevy's Zebras Landscape of Fear }
\label{cs1}

\begin{figure}[htbp]
    \centering
    \begin{subfigure}[b]{0.48\textwidth}
        \centering
        \includegraphics[width=\textwidth]{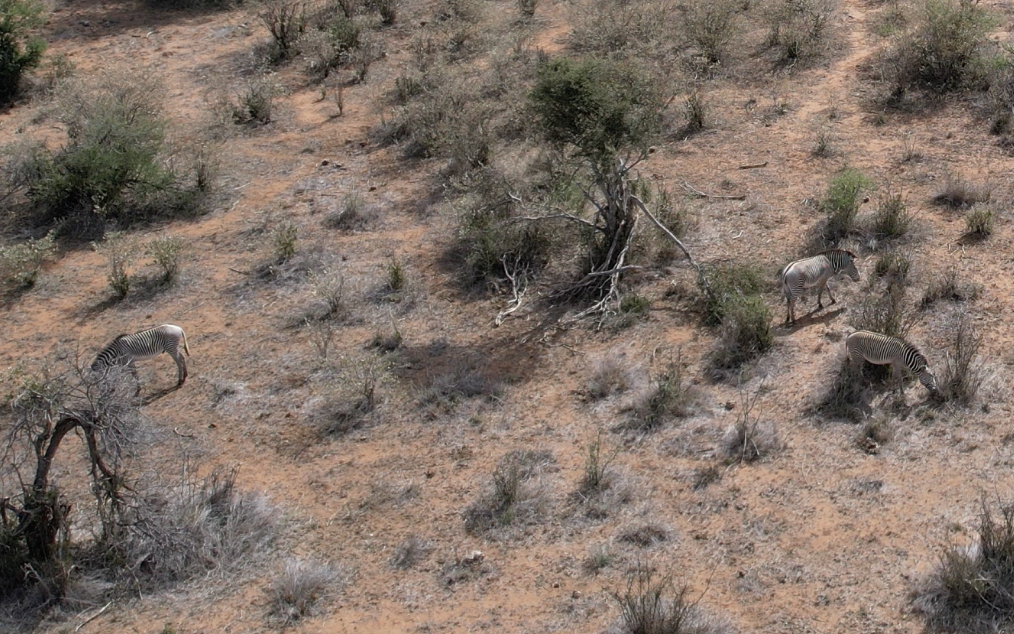}
        \caption{Example of closed habitat with dense vegetation (Session 4 from Tab. \ref{tab:habitat_data}), \citep{kline_kabr_telemetry}.}
        \label{fig:photo1}
    \end{subfigure}
    \hfill
    \begin{subfigure}[b]{0.48\textwidth}
        \centering
        \includegraphics[width=\textwidth]{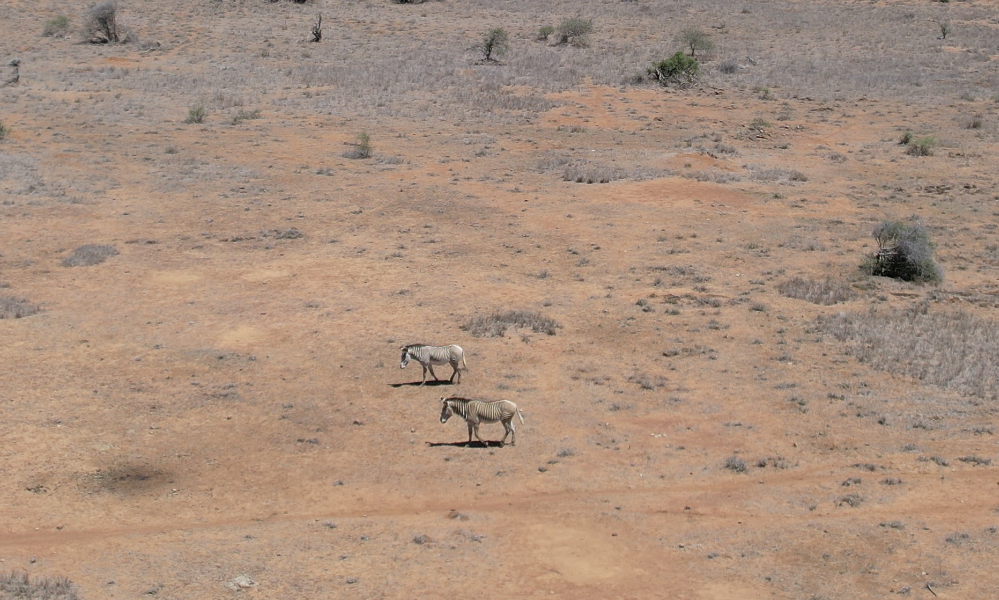}
        \caption{Example of open habitat with with little vegetation (Session 5 from Tab. \ref{tab:habitat_data}), \citep{KABR_worked_example}.}
        \label{fig:photo2}
    \end{subfigure}
    \caption{Example of Closed versus Open habitats analyzed for Case Study 1: \textit{Grevy's Zebras Landscape of Fear}.}
    \label{fig:openvsclosed}
\end{figure}

We analyze anti-predator strategies in Grevy's zebras using 83 mini-scenes \citep{kabr_telemetry_dataset, KABR_worked_example}. We selected mini-scenes from six representative sessions, containing a mix of herd sizes and habitat types (Tab. \ref{tab:habitat_data}). To ensure sufficiently long behavior sequences for analysis, we selected mini-scenes $\geq 60$ seconds. We classify the mini-scenes by habitat type based on vegetation density by visually examining the mini-scene footage to classify habitat type as open (little to no vegetation) or closed (dense vegetation). Examples of habitat categories are illustrated in Fig. \ref{fig:openvsclosed}. The median herd size is three individuals; herds $\le 3$ are categorized as small, herds $>3$ are large. For each mini-scene, the time budget was calculated as a proportion of \textit{Walk}, \textit{Head Up}, \textit{Graze}, and \textit{Other}. We use \textit{Head Up} behaviors to indicate instances of vigilance, where animals survey their habitat for potential risks. 

\begin{figure}
    \centering
    \includegraphics[width=1\linewidth]{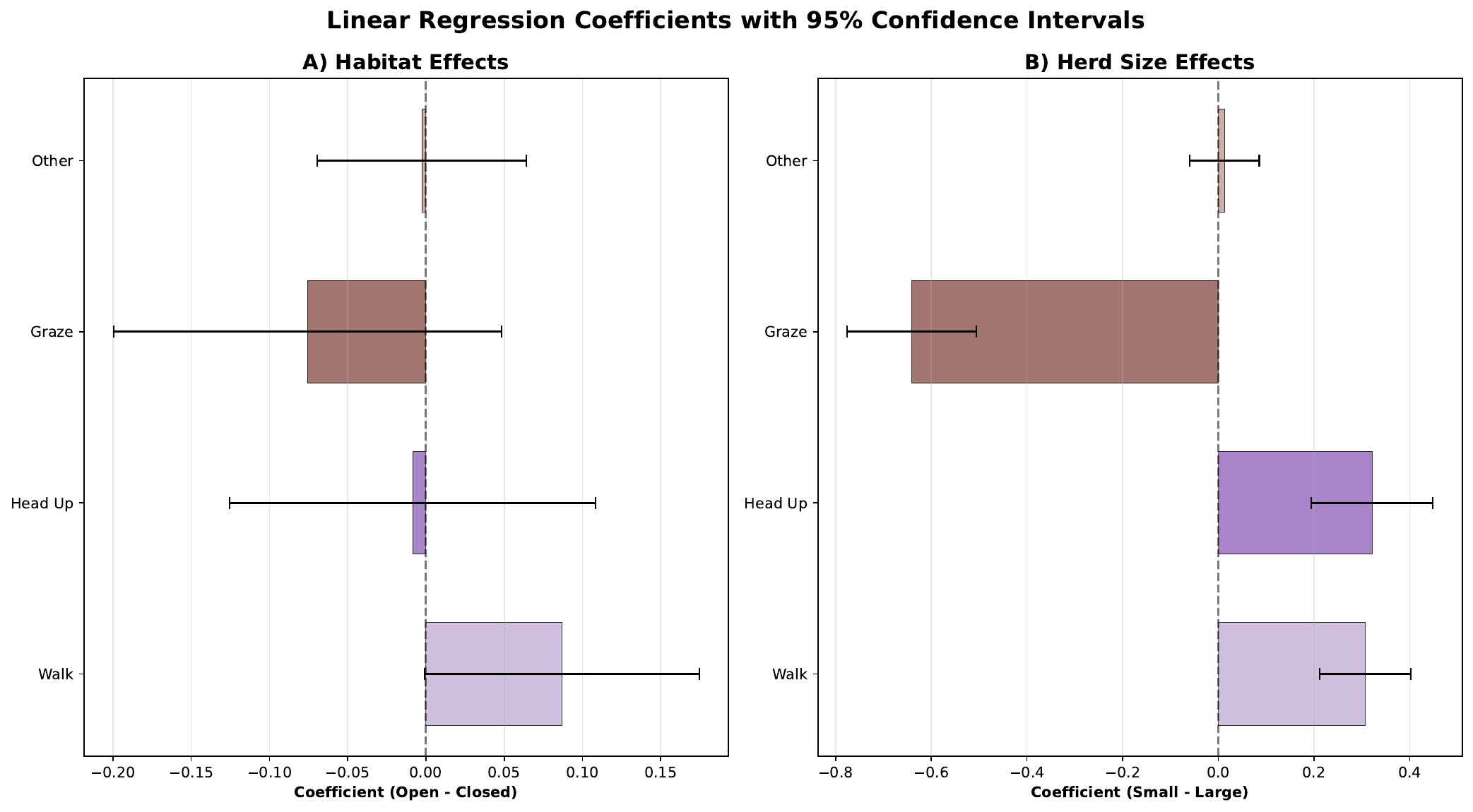}
    \caption{Linear regression coefficients with 95\% confidence intervals showing the effect of habitat (Open vs Closed) and herd size (Small vs Large) on four behaviors, analyzed for Case Study 1: \textit{Grevy's Zebras Landscape of Fear}. Larger herds allow Grevy’s zebras to spend more time grazing and less time scanning or walking, confirming the classic group -- vigilance trade‑off. The habitat panels show that any raw differences among closed, mixed, and open bush are largely explained by the bigger groups that congregate in dense cover -- once herd size is accounted for, habitat adds little additional effect.}
    \label{fig:linear_regression}
\end{figure}

We conducted linear regression analyses to examine the effects of habitat type (closed vs. open) and herd size (large vs. small) on the four behavioral categories, summarized in Tab. \ref{tab:regression_summary}. Categorical predictor variables were dummy coded with closed habitat and large herds serving as reference categories. For each behavior, we fitted separate multiple linear regression models using ordinary least squares estimation. Model assumptions were verified through residual analysis, and statistical significance was assessed using t-tests with $\alpha = 0.05$. We calculated 95\% confidence intervals for all coefficients and quantified effect sizes using Cohen's $f^2$, visualized in Fig. \ref{fig:linear_regression}. We also tested for potential interaction effects between habitat and herd size but found no significant interactions, confirming that the effects of these variables are additive rather than synergistic.
The linear regression models revealed substantial variation in how social and environmental context influence different behaviors, with model performance ($R^2$) ranging from negligible for \textit{Other} behaviors ($R^2 = 0.002$) to strong for \textit{Graze} ($R^2 = 0.562$). Herd size emerged as the dominant predictor across all behaviors, showing significant effects for \textit{Walk} ($\beta = 0.307, p < 0.001$), \textit{Head Up} ($\beta = 0.321, p < 0.001$), and \textit{Graze} ($\beta = -0.641, p < 0.001$), with small herds consistently showing higher activity levels except for \textit{Graze}. Habitat effects were generally weaker and non-significant, with only marginal significance for walking behavior ($\beta = 0.087, p = 0.052$). The strongest and most predictable response was \textit{Graze}, which decreased substantially in small compared to large herds, while vigilance and locomotory behaviors increased.

\begin{table*}[t]
\centering
\begin{tabular}{lccc}
\toprule
\textbf{Behavior} & \textbf{$R^2$} & \textbf{Habitat Effect (Open vs Closed)} & \textbf{Herd Size Effect (Small vs Large)} \\
\midrule
Walk & 0.402 *** & 0.087† & 0.307 *** \\
Head Up & 0.250 *** & -0.008 & 0.321 *** \\
Graze & 0.562 *** & -0.076 & -0.641 *** \\
Other & 0.002     & -0.003 & 0.013    \\
\bottomrule \\ 
\end{tabular}
\caption{Summary of linear regression results examining habitat and herd size effects on Grevy's zebra behaviors. The reported coefficients represent a change in the proportion of time spent on behavior. Small herds significantly increased walking, vigilance (head up), and decreased grazing compared to large herds (all p < 0.001), while habitat type showed minimal effects on behavior, with herd size explaining 25-56\% of behavioral variation depending on the activity.\\ 
\footnotesize Significance levels: $*** \ p < 0.001; ** \ p < 0.01; * \ p < 0.05; † \ p < 0.10$. Reference categories: Closed habitat, Large herds.}
\label{tab:regression_summary}
\end{table*}

\newpage
\subsubsection{Case Study 2: Zebra State Shifts}
\label{cs2}

\begin{wrapfigure}{r}{0.4\textwidth}
    \scriptsize
    \centering
    \includegraphics[width=1\linewidth]{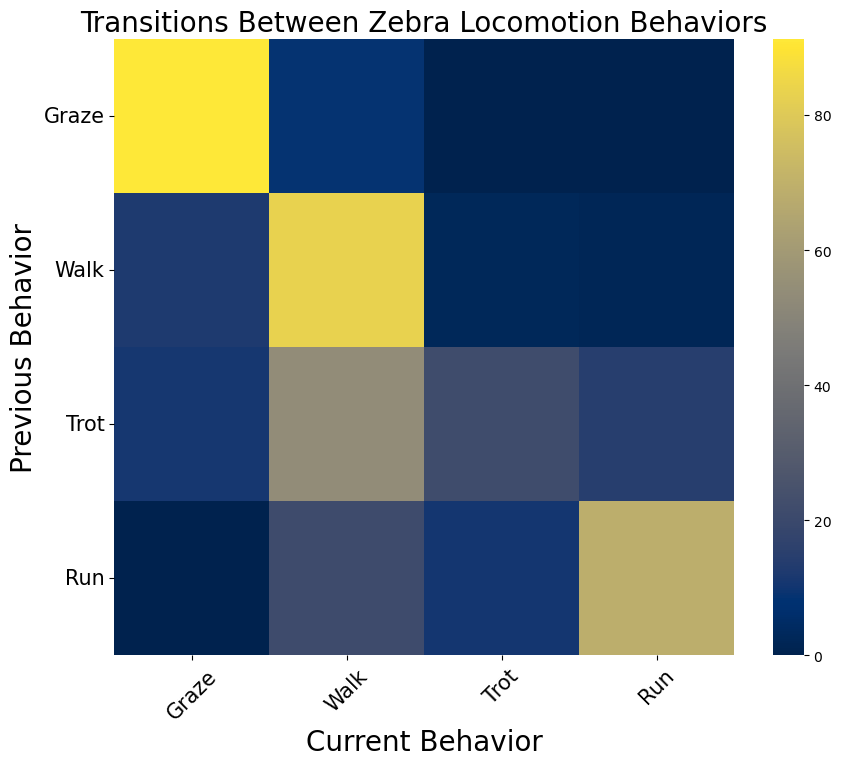}
    \caption{Transition matrix for zebra locomotion behaviors, ordered by least to greatest velocity, for Case Study 2: Z\textit{ebra State Shifts}.}
    \label{fig:behaviortransitions}
\end{wrapfigure}

We analyzed 748 mini-scenes \citep{kabr_telemetry_dataset}. The mini-scenes total 2 hours 30 minutes, containing zebra locomotion behaviors of \textit{Graze}, \textit{Walk}, \textit{Trot}, and \textit{Run}. 
We quantify the likelihood of zebras transitioning between behaviors (Fig.~\ref{fig:behaviortransitions}) by normalizing the transitions by the count of occurrences, down-sampled every 10 seconds. Individuals are most likely to remain in the same behavioral state, with high self-transition probabilities for \textit{Graze} (91.1\%), \textit{Walk} (83.0\%), and \textit{Run} (68.4\%). Previous instances of \textit{Trot} were most likely to transition to \textit{Walk} (53.4\%). Transitions from relaxed states to more alert behaviors are also evident: for instance, individuals in a \textit{Walk} state transition to \textit{Trot} 2.6\% of the time and to \textit{Run} 2.0\% of the time, while \textit{Trot} transitions to \textit{Run} with a relatively high probability of 14.3\%. These transition dynamics provide insight into how animals escalate movement in response to internal or external stimuli and offer a foundation for modeling behavioral state changes. Understanding these patterns is critical for developing behavior-adaptive autonomous drones that can anticipate transitions, such as from grazing to fleeing, and adjust drone flight paths accordingly to minimize disturbance while maximizing data capture.

\subsubsection{Case Study 3: Cross-Species Encounters}
\label{cs3}

\begin{figure}
    \centering
    \includegraphics[width=1\linewidth]{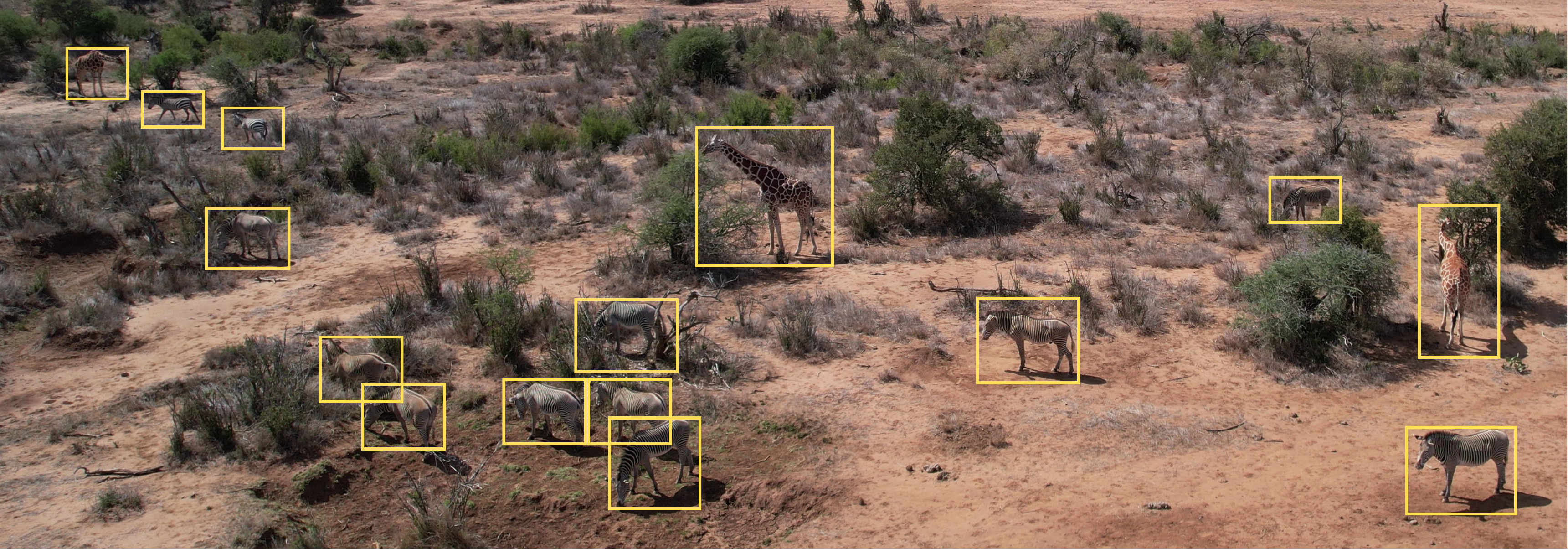}
    \caption{Image from mini-scene of mixed species herd consisting of 11 Grevy's zebras (center), 2 Plains zebras (left), and 3 reticulated giraffes (left, center, right) analyzed for Case Study 3:\textit{ Cross-Species Encounters} \citep{KABR_worked_example}. }
    \label{fig:mixedherd}
\end{figure}

A mixed species herd of Grevy's zebras, Plains zebras, and giraffes was observed grazing together, illustrated in Fig. \ref{fig:mixedherd}) \citep{KABR_worked_example}. From the drone videos, 157 mini-scenes were generated and automatically labeled with behaviors using the KABR X3D model \citep{kabr_x3d_model}. To assess patterns of spatial proximity, we computed normalized bounding box overlap counts between individuals of the same and different species (Tab. \ref{tab:normalized_overlap}). Normalization was performed by dividing the total number of overlaps by the number of possible pairs for each species combination. Both zebra species showed higher overlap compared to giraffes. Grevy's zebras exhibited a high normalized overlap rate (87.93 overlaps per pair), suggesting frequent and sustained close proximity among group members. Plains zebras showed the highest normalized value (93.00), though this reflects a single dyad and should be interpreted with caution due to the small sample size. Giraffes had a moderate normalized overlap rate (26.00), consistent with a more spatially dispersed grouping structure. Cross-species overlaps were rare. The normalized overlap between Grevy's and plains zebras was low (1.27), and no overlaps were observed between giraffes and zebras (0.00), indicating strong spatial segregation between species in the observed dataset. Overall, these patterns suggest that intra-species proximity is most prominent in Grevy's zebras, with limited interaction between species.

\begin{table}[t]
\centering
\begin{tabular}{lrrrr}
\toprule
\textbf{Species Pair} & \textbf{Overlap Count} & \textbf{Possible Pairs} & \textbf{Normalized Overlap} \\
\midrule
Grevy's Zebra -- Grevy's Zebra   & 4836 & 55  & 87.93 \\
Plains Zebra -- Plains Zebra   & 93   & 1   & 93.00 \\
Giraffe -- Giraffe             & 78   & 3   & 26.00 \\
Grevy's Zebra -- Plains Zebra   & 28   & 22  & 1.27 \\
Giraffe -- Plains Zebra        & 0    & 6   & 0.00 \\
Giraffe-- Grevy's Zebra         & 0    & 33   & 0.00 \\
\bottomrule \\
\end{tabular}
\caption{Bounding box overlap counts and normalized values for intra- and inter-species interactions for Case Study 3: \textit{Cross-Species Encounters}.. Normalized values represent average overlaps per possible pair. Same species: $n(n-1)/2$ possible pairs. Cross-species: $n_1 \times n_2$. 11 Grevy's zebras, 3 reticulated giraffes, and 2 Plains zebras.}
\label{tab:normalized_overlap}
\end{table}

\section{Discussion}
Our results demonstrate that drone-based behavioral observation, augmented by machine learning, offers a scalable, non-invasive, and ecologically valid method for studying animal behavior in natural environments. In Section \ref{video}, we discuss our proposed hybrid semi-automated approach for video processing to balance scalability with accuracy. In Section \ref{drone}, we discuss how drone-based observation overcomes the fundamental limitations of traditional ground-based methods by providing simultaneous focal sampling of multiple individuals with superior temporal resolution and fewer observation gaps, bridging the trade-off between behavioral detail and group-level context. In Section \ref{casestudies}, we discuss the ecological insights derived from the case studies, including species-specific vigilance patterns in Grevy's zebras, structured behavioral state transitions, and spatial segregation in mixed-species herds. Together, these findings suggest that drone-based automated behavioral analysis provides a valuable complement to traditional methods, offering researchers new opportunities to study animal behavior at larger scales while acknowledging the ongoing need for methodological refinement and human oversight.

\subsection{Video Processing Pipeline}
\label{video}
Our pipeline processes drone video to enable scalable, semi-automated analysis of animal behavior in ecological field studies. It integrates automated machine learning with targeted manual annotation, balancing scalability with accuracy. While fully manual annotation yields highly detailed behavioral records, the time investment becomes prohibitive at scale. Instead, we propose a hybrid workflow in which machine learning models handle dominant behaviors, while human annotators validate edge cases and review rare or ambiguous behaviors.

We designed our pipeline to be modular, allowing users to easily integrate emerging machine learning models. Our X3D model performed well in classifying dominant behaviors such as \textit{Walk, Graze}, and \textit{Head} Up, achieving over 95\% mean average precision (mAP) and a 91.4\% F1 score for these classes. However, its performance declined for rare or visually ambiguous behaviors, such as \textit{Run} and \textit{Auto-groom}. This outcome reflects the long-tail problem in machine learning, where rare behaviors suffer from limited training examples and, as a result, lower predictive accuracy. Despite this limitation, the X3D model is sufficient for identifying broad behavioral trends with a per-instance (micro) average accuracy of 86.7\% \citep{kholiavchenko2024kabra}. It is also sufficient for automatically estimating coarse-grained time budgets, such as those used for Case Study 1 (Section \ref{cs1}). Automated methods struggle with high-resolution or rare-behavior analyses, motivating the need for human-in-the-loop processing pipelines to capture these instances.
When integrating new models into the pipeline, we recommend setting model confidence thresholds (e.g., $>0.8$) to enhance precision, with the understanding that this may reduce recall. For studies requiring high accuracy or detailed behavioral timing, manual review remains essential.

Manual annotation provides fine-grained, accurate behavioral labels but is labor-intensive. Based on our annotation rate model ($T_{\text{annotation}} = 1.5nt$ for $n$ individuals over $t$ seconds), a 10-minute video with three individuals requires approximately 15 minutes of annotation time per animal, for a total of 45 minutes to process the entire scene. This annotation burden scales linearly with group size and video length, making fully manual approaches impractical for large datasets. Our project included 969 annotated mini-scenes, which would have required hundreds of hours of effort without automation. The semi-automated workflow we developed offers a practical compromise between scale and precision. Automated classification handles the bulk of labeling for frequent behaviors, while human annotators focus on reviewing edge cases, validating model outputs, and labeling rare behaviors. To ensure reliability across annotators and datasets, we emphasize the importance of standardized annotation protocols and regular calibration sessions to maintain inter-annotator consistency.

The effectiveness of the pipeline depends not only on model accuracy but also on data quality and experimental design. Common technical challenges include object detection failures when animals are obscured by vegetation or overlapping conspecifics, leading to \textit{Out of Sight} segments. In addition, video resolution limits can constrain the identification of individuals and subtle behaviors, particularly at greater distances or flight altitudes. Careful flight planning is essential to optimize resolution and field of view for behavioral monitoring. Finally, behavior definitions must be operationalized clearly to reduce inconsistencies, especially for transitional or compound behaviors.

Our results demonstrate that semi-automated analysis offers a powerful path forward for behavioral ecology, enabling analysis at scales that would otherwise be infeasible. While current model performance imposes constraints--particularly for rare behaviors--hybrid approaches that combine machine learning with human oversight provide a scalable and accurate framework. Future improvements in model architecture, training data, and multimodal integration will continue to expand the ecological questions that automated pipelines like \texttt{kabr-tools} can address.

\subsection{Drone-based Behavioral Observations}
\label{drone}
Our systematic comparison of behavioral sampling methods reveals distinct advantages and limitations of traditional ground-based scan and focal sampling. Focal sampling consistently captures more detailed behavioral information than scan sampling, detecting rare behaviors (\textit{Auto-groom, Fight}) that are missed by instantaneous scan observations. However, scan sampling provides the critical advantage of capturing behavioral data from multiple individuals under identical environmental and social contexts, eliminating the temporal confounds inherent in sequential focal observations. Drone-based observations bridge these methodological gaps by fine-grained focal sampling of multiple individuals under identical conditions. Compared to ground-based methods, drones experienced fewer observation gaps, captured more behavioral transitions, and provided richer behavioral sequences, especially for brief behaviors like vigilance. We did not conduct simultaneous drone missions and scan sampling sessions. However, this would be a valuable contribution for future work, particularly in comparing tradeoffs in cost and labor for each methodology and granularity of behavior data captured. The superior temporal resolution of drone annotations stems from two key factors: (1) the aerial perspective minimizes visual obstructions, and (2) post-hoc video analysis allows annotators to pause and rewind footage, ensuring precise alignment of behavioral transitions with video frames. This precision advantage is particularly evident for brief behaviors like vigilance responses, where ground observers may introduce recording delays that artificially inflate behavior durations.

\subsection{Ecological Case Studies}
\label{casestudies}
To demonstrate the capabilities of our \texttt{kabr-tools} pipeline, we applied it to three ecological case studies: anti-predator behavior, behavioral state transitions, and group structure in mixed-species herds (Section \ref{discussion_cs}). These case studies underscore the types of ecological insights that drone-based automated analysis enables at scale (Section \ref{strengths}).

\subsubsection{Case Studies Findings}
\textbf{Case Study 1: Grevy's Zebra Landscape of Fear.}
\label{discussion_cs}
We analyzed 86 mini-scenes of Grevy's herds ranging from two to seven individuals across open and closed habitats. We found that individual vigilance decreased significantly with increasing group size, consistent with the group-vigilance trade-off \citep{rubenstein1986ecology}. Notably, habitat type had no additional effect on vigilance after controlling for group size--unlike in plains zebras, where habitat remains a significant factor \citep{Chen_2021}. This novel finding reveals a fundamental difference in risk assessment by zebra species. The tradeoff between survival and nutrition significantly impacts future reproduction \citep{rubenstein1986ecology}. 
The difference in vigilance trade-offs between Grevy's and plains zebras likely stems from social structure, specifically the differences in the fluidity of each species' social system \citep{rubenstein2010ecology}. Plains zebras exhibit more rigid groupings, while Grevy's zebras form flexible, fission-fusion groups, enabling fine-tuned responses to risk through dynamic group size adjustment. While plains zebras live in fixed closed membership groups that can join and leave herds, Grevy's zebra individuals are freer to move in and out of herds as conditions change, so responding to changes in risk can happen more quickly, making individuals less risk-averse than individual plains zebras, their close evolutionary kin.
However, it should be noted that these conclusions are drawn from five sampling sessions representing 20 individuals. Further data collection should be conducted to explore this relationship further.
Grevy's zebras are endangered, with less than two thousand individuals remaining in the wild \citep{rubenstein2016equus}, so understanding factors driving their future reproduction is vital for protecting and preserving these at-risk populations.

\textbf{Case Study 2: Zebra State Shifts.}
We utilized 748 mini-scenes of behavioral sequences to construct a transition matrix between locomotor states (\textit{Graze}, \textit{Walk}, \textit{Trot}, \textit{Run}). These revealed strong behavioral inertia and structured escalation patterns, where transitions from relaxed states (e.g., \textit{Graze}) to alert states (e.g., \textit{Run}) occurred in measurable steps. This structure provides predictive value for behavior-adaptive autonomous sensing, offering a potential feedback loop between real-time drone behavior and animal response.

\textbf{Case Study 3: Cross-Species Encounters.}
We examined spatial interactions in a mixed-species herd using bounding box overlap metrics. Intra-species proximity was strongest in both Grevy's and Plains zebras, while inter-species interactions were rare despite shared space. These findings indicate spatial segregation across species and demonstrate the utility of drone imagery for quantifying social cohesion and species-level group structure in dynamic field settings.

\subsubsection{Implications of Case Studies}
\label{strengths}
Across the three case studies, our framework demonstrated three key strengths: scale, simultaneity, and standardization. We analyzed 969 mini-scenes -- far beyond what traditional methods could feasibly capture or annotate. This scale enables robust statistical inference on behavior and interactions across ecological gradients. Drone-based observation captures all individuals in a group under identical environmental and social conditions, providing superior simultaneity. This eliminates temporal confounds common in sequential focal sampling and was essential for detecting group-level behavioral patterns. Finally, automated labeling ensures consistent behavioral definitions across observations, minimizing observer bias and facilitating standardized comparisons across species and sites.

While our results reveal important ecological insights, they also highlight the current limitations of automated behavioral classification. With a classification accuracy of approximately 91\% for the three behavior classes used for \textit{Grevy's Landscape of Fear} case study, we can be reasonably certain the automatically generated time budgets are accurate. However, the automatic recognition of rare behaviors and precise timing remains challenging. Despite this, the strength of statistical patterns emerging from large sample sizes suggests that automated annotation is already a powerful tool for broad ecological inference. Future work will focus on improving behavior classification through model fine-tuning, incorporating multimodal data, and human-in-the-loop workflows to expand the range of ecological questions that can be addressed. Ultimately, this work demonstrates that drone-based automated analysis is not merely a faster alternative to manual methods, but a transformative tool enabling new modes of behavioral and ecological inquiry.

\section{Conclusion}
Our work demonstrates that the integration of drone technology, computer vision, and machine learning creates new possibilities for ecological research while highlighting the importance of rigorous validation and performance assessment. The success of this approach depends on continued collaboration between ecological and computational disciplines, with particular attention to developing tools that serve genuine research needs while maintaining the highest standards of scientific rigor and animal welfare. The \texttt{kabr-tools} framework and associated dataset provide a foundation for advancing automated wildlife behavioral analysis. Realizing the full potential of these approaches requires sustained effort in model development, standardization, and building collaborative frameworks, rooted in application-driven machine learning \citep{rolnick2024application}. The future of ecological informatics lies in creating seamless integration between field observations, automated data collection, and predictive modeling to support evidence-based conservation and management decisions.

\section*{Acknowledgments}
This work was supported by the Imageomics Institute, which is funded by the US National Science Foundation's Harnessing the Data Revolution (HDR) program under Award 2118240 (Imageomics: A New Frontier of Biological Information Powered by Knowledge-Guided Machine Learning). Additional support was provided by the AI Institute for Intelligent Cyberinfrastructure with Computational Learning in the Environment (ICICLE), funded by the US National Science Foundation under Award 2112606.  We thank the Mpala Research Centre for their support in collecting the dataset.

\section{Data Availability Statement}
All code, annotated datasets, and documentation are publicly available at \url{https://github.com/Imageomics/kabr-tools}. Documentation in Github includes annotation guidelines and best practices for managing large-scale ecological video datasets. The data used for analysis is available on HuggingFace and includes \href{https://huggingface.co/datasets/imageomics/kabr-full-video}{full-length videos} \citep{KABR_Raw_Videos}, \href{https://huggingface.co/datasets/imageomics/kabr-worked-examples}{worked examples} \citep{KABR_worked_example}, and \href{https://huggingface.co/datasets/imageomics/kabr-methodology}{methodology comparison} \citep{kline2025kabr-tools-methodology}.

The data was collected at the Mpala Research Centre in Kenya, in accordance with Research License No. NACOSTI/P/22/18214. The data collection protocol adhered strictly to the guidelines set forth by the Institutional Animal Care and Use Committee under permission No. IACUC 1835F.

\bibliographystyle{plainnat}  
\bibliography{references}  

\begin{thebibliography}{44}
\providecommand{\natexlab}[1]{#1}
\providecommand{\url}[1]{\texttt{#1}}
\expandafter\ifx\csname urlstyle\endcsname\relax
  \providecommand{\doi}[1]{doi: #1}\else
  \providecommand{\doi}{doi: \begingroup \urlstyle{rm}\Url}\fi

\bibitem[Afridi et~al.(2024)Afridi, Hlebowicz, Cawthorne, and Lundquist]{afridi2024noise}
Saadia Afridi, Kasper Hlebowicz, Dylan Cawthorne, and Ulrik Lundquist.
\newblock Unveiling the impact of drone noise on wildlife: A crucial research imperative.
\newblock pages 1409--1416, 06 2024.
\newblock \doi{10.1109/ICUAS60882.2024.10557094}.

\bibitem[Altmann(1974)]{Altmann_1974}
Jeanne Altmann.
\newblock Observational study of behavior: Sampling methods.
\newblock \emph{Behaviour}, 49\penalty0 (3–4):\penalty0 227–266, January 1974.
\newblock ISSN 0005-7959, 1568-539X.
\newblock \doi{10.1163/156853974X00534}.

\bibitem[Amato et~al.(2013)Amato, Van~Belle, and Wilkinson]{Amato_Van_Belle_Wilkinson_2013}
Katherine~R. Amato, Sarie Van~Belle, and Brianna Wilkinson.
\newblock A comparison of scan and focal sampling for the description of wild primate activity, diet and intragroup spatial relationships.
\newblock \emph{Folia Primatologica; International Journal of Primatology}, 84\penalty0 (2):\penalty0 87–101, 2013.
\newblock ISSN 1421-9980.
\newblock \doi{10.1159/000348305}.

\bibitem[Anderson and Perona(2014)]{Anderson_Perona_2014}
David~J. Anderson and Pietro Perona.
\newblock Toward a science of computational ethology.
\newblock \emph{Neuron}, 84\penalty0 (1):\penalty0 18–31, October 2014.
\newblock ISSN 0896-6273.
\newblock \doi{10.1016/j.neuron.2014.09.005}.

\bibitem[Axford et~al.(2024)Axford, Sohel, Vanderklift, and Hodgson]{axford2024collectively}
Daniel Axford, Ferdous Sohel, Mathew Vanderklift, and Amanda Hodgson.
\newblock Collectively advancing deep learning for animal detection in drone imagery: Successes, challenges, and research gaps.
\newblock \emph{Ecological informatics}, page 102842, 2024.

\bibitem[Bateson and Martin(2021)]{bateson}
M.~Bateson and P.~Martin.
\newblock \emph{Measuring behaviour: an introductory guide}.
\newblock Cambridge university press, 2021.

\bibitem[Besson et~al.(2022)Besson, Alison, Bjerge, Gorochowski, H{\o}ye, Jucker, Mann, and Clements]{besson2022towards}
Marc Besson, Jamie Alison, Kim Bjerge, Thomas~E Gorochowski, Toke~T H{\o}ye, Tommaso Jucker, Hjalte~MR Mann, and Christopher~F Clements.
\newblock Towards the fully automated monitoring of ecological communities.
\newblock \emph{Ecology Letters}, 25\penalty0 (12):\penalty0 2753--2775, 2022.

\bibitem[Carreira and Zisserman(2017)]{carreira2017quo}
Joao Carreira and Andrew Zisserman.
\newblock Quo vadis, action recognition? a new model and the kinetics dataset.
\newblock In \emph{proceedings of the IEEE Conference on Computer Vision and Pattern Recognition}, pages 6299--6308, 2017.

\bibitem[Chan et~al.(2024)Chan, Putra, Schupp, K{\"o}chling, Stra{\ss}heim, Renner, Schroeder, Pearse, Nakagawa, Burke, et~al.]{chan2024yolo}
Alex Hoi~Hang Chan, Prasetia Putra, Harald Schupp, Johanna K{\"o}chling, Jana Stra{\ss}heim, Britta Renner, Julia Schroeder, William~D Pearse, Shinichi Nakagawa, Terry Burke, et~al.
\newblock Yolo-behaviour: A simple, flexible framework to automatically quantify animal behaviours from videos.
\newblock \emph{bioRxiv}, pages 2024--08, 2024.

\bibitem[Chen et~al.(2021)Chen, Reperant, Fischhoff, and Rubenstein]{Chen_2021}
Anping Chen, Leslie Reperant, Ilya~R. Fischhoff, and Daniel~I. Rubenstein.
\newblock Increased vigilance of plains zebras (equus quagga) in response to more bush coverage in a kenyan savanna.
\newblock \emph{Climate Change Ecology}, 1:\penalty0 100001, July 2021.
\newblock ISSN 2666-9005.
\newblock \doi{10.1016/j.ecochg.2021.100001}.

\bibitem[Corcoran et~al.(2021)Corcoran, Winsen, Sudholz, and Hamilton]{Corcoran_Winsen_Sudholz_Hamilton_2021}
Evangeline Corcoran, Megan Winsen, Ashlee Sudholz, and Grant Hamilton.
\newblock Automated detection of wildlife using drones: Synthesis, opportunities and constraints.
\newblock \emph{Methods in Ecology and Evolution}, 12\penalty0 (6):\penalty0 1103–1114, 2021.
\newblock ISSN 2041-210X.
\newblock \doi{10.1111/2041-210X.13581}.

\bibitem[{CVAT.ai}(2023)]{CVAT}
{CVAT.ai}.
\newblock {Computer Vision Annotation Tool (CVAT)}, November 2023.
\newblock URL \url{https://github.com/cvat-ai/cvat}.

\bibitem[Duporge et~al.(2021)Duporge, Spiegel, Thomson, Chapman, Lamberth, Pond, Macdonald, Wang, and Klinck]{duporge2021determination}
Isla Duporge, Marcus~P Spiegel, Eleanor~R Thomson, Tatiana Chapman, Curt Lamberth, Caroline Pond, David~W Macdonald, Tiejun Wang, and Holger Klinck.
\newblock Determination of optimal flight altitude to minimise acoustic drone disturbance to wildlife using species audiograms.
\newblock \emph{Methods in Ecology and Evolution}, 12\penalty0 (11):\penalty0 2196--2207, 2021.

\bibitem[Duporge et~al.(2025)Duporge, Kholiavchenko, Harel, Wolf, Rubenstein, Crofoot, Berger-Wolf, Lee, Barreau, Kline, et~al.]{duporge2025baboonland}
Isla Duporge, Maksim Kholiavchenko, Roi Harel, Scott Wolf, Daniel~I Rubenstein, Margaret~C Crofoot, Tanya Berger-Wolf, Stephen~J Lee, Julie Barreau, Jenna Kline, et~al.
\newblock Baboonland dataset: Tracking primates in the wild and automating behaviour recognition from drone videos: I. duporge et al.
\newblock \emph{International Journal of Computer Vision}, pages 1--12, 2025.

\bibitem[Elhorst et~al.(2025)Elhorst, Syposz, and Wojczulanis-Jakubas]{elhorst2025behave}
Reinoud Elhorst, Martyna Syposz, and Katarzyna Wojczulanis-Jakubas.
\newblock Behave-facilitating behaviour coding from videos with ai-detected animals.
\newblock \emph{Ecological Informatics}, 87:\penalty0 103106, 2025.

\bibitem[Feichtenhofer(2020)]{feichtenhofer2020x3d}
Christoph Feichtenhofer.
\newblock X3d: Expanding architectures for efficient video recognition.
\newblock In \emph{Proceedings of the IEEE/CVF conference on computer vision and pattern recognition}, pages 203--213, 2020.

\bibitem[Feichtenhofer et~al.(2019)Feichtenhofer, Fan, Malik, and He]{slowfast_2019}
Christoph Feichtenhofer, Haoqi Fan, Jitendra Malik, and Kaiming He.
\newblock Slowfast networks for video recognition.
\newblock In \emph{2019 IEEE/CVF International Conference on Computer Vision (ICCV)}, page 6201–6210, Seoul, Korea (South), October 2019. IEEE.
\newblock ISBN 978-1-72814-803-8.
\newblock \doi{10.1109/ICCV.2019.00630}.
\newblock URL \url{https://ieeexplore.ieee.org/document/9008780/}.

\bibitem[Hughey et~al.(2018)Hughey, Hein, Strandburg-Peshkin, and Jensen]{hughey2018challenges}
Lacey~F Hughey, Andrew~M Hein, Ariana Strandburg-Peshkin, and Frants~H Jensen.
\newblock Challenges and solutions for studying collective animal behaviour in the wild.
\newblock \emph{Philosophical Transactions of the Royal Society B: Biological Sciences}, 373\penalty0 (1746):\penalty0 20170005, 2018.

\bibitem[Jocher et~al.(2023)Jocher, Chaurasia, and Qiu]{yolo_2023}
Glenn Jocher, Ayush Chaurasia, and Jing Qiu.
\newblock Yolo by ultralytics, January 2023.
\newblock URL \url{https://github.com/ultralytics/ultralytics}.

\bibitem[Kholiavchenko et~al.(2024{\natexlab{a}})Kholiavchenko, Kline, Kukushkin, Brookes, Stevens, Duporge, Sheets, Babu, Banerji, Campolongo, Thompson, Tiel, Miliko, Bessa, Mirmehdi, Schmid, Berger-Wolf, Rubenstein, Burghardt, and Stewart]{kabrdeepdive}
Maksim Kholiavchenko, Jenna Kline, Maksim Kukushkin, Otto Brookes, Sam Stevens, Isla Duporge, Alec Sheets, Reshma~R. Babu, Namrata Banerji, Elizabeth Campolongo, Matthew Thompson, Nina~Van Tiel, Jackson Miliko, Eduardo Bessa, Majid Mirmehdi, Thomas Schmid, Tanya Berger-Wolf, Daniel~I. Rubenstein, Tilo Burghardt, and Charles~V. Stewart.
\newblock Deep dive into kabr: a dataset for understanding ungulate behavior from in-situ drone video.
\newblock \emph{Multimedia Tools and Applications}, December 2024{\natexlab{a}}.
\newblock ISSN 1573-7721.
\newblock \doi{10.1007/s11042-024-20512-4}.
\newblock URL \url{https://doi.org/10.1007/s11042-024-20512-4}.

\bibitem[Kholiavchenko et~al.(2024{\natexlab{b}})Kholiavchenko, Kline, Ramirez, Stevens, Sheets, Babu, Banerji, Campolongo, Thompson, Van~Tiel, Miliko, Bessa, Duporge, Berger-Wolf, Rubenstein, and Stewart]{kholiavchenko2024kabra}
Maksim Kholiavchenko, Jenna Kline, Michelle Ramirez, Sam Stevens, Alec Sheets, Reshma Babu, Namrata Banerji, Elizabeth Campolongo, Matthew Thompson, Nina Van~Tiel, Jackson Miliko, Eduardo Bessa, Isla Duporge, Tanya Berger-Wolf, Daniel Rubenstein, and Charles Stewart.
\newblock Kabr: In-situ dataset for kenyan animal behavior recognition from drone videos.
\newblock In \emph{Proceedings of the IEEE/CVF Winter Conference on Applications of Computer Vision}, pages 31--40, January 2024{\natexlab{b}}.

\bibitem[Kholiavchenko et~al.(2024{\natexlab{c}})Kholiavchenko, Kukushkin, Brookes, Kline, Stevens, Duporge, Sheets, Babu, Banerji, Campolongo, Thompson, Van~Tiel, Miliko, Mirmehdi, Schmid, Berger-Wolf, Rubenstein, Burghardt, and Stewart]{kabr_x3d_model}
Maksim Kholiavchenko, Maksim Kukushkin, Otto Brookes, Jenna Kline, Sam Stevens, Isla Duporge, Alec Sheets, Reshma~R. Babu, Namrata Banerji, Elizabeth Campolongo, Matthew Thompson, Nina Van~Tiel, Jackson Miliko, Eduardo~Bessa Mirmehdi, Thomas Schmid, Tanya Berger-Wolf, Daniel~I. Rubenstein, Tilo Burghardt, and Charles~V. Stewart.
\newblock Kabr model, 2024{\natexlab{c}}.
\newblock URL \url{https://huggingface.co/imageomics/x3d-kabr-kinetics}.

\bibitem[Kline et~al.(2024{\natexlab{a}})Kline, Kholiavchenko, Berger-Wolf, Stewart, and Stewart]{kline_kabr_telemetry}
Jenna Kline, Maksim Kholiavchenko, Tanya Berger-Wolf, Charles~V. Stewart, and Christopher Stewart.
\newblock Integrating biological data into autonomous remote sensing systems for in situ imageomics: A case study for kenyan animal behavior sensing with unmanned aerial vehicles (uavs).
\newblock In \emph{Proceedings of the First Workshop on Imageomics: Discovering Biological Knowledge from Images using AI, held as part of AAAI 24}, 2024{\natexlab{a}}.

\bibitem[Kline et~al.(2024{\natexlab{b}})Kline, Kholiavchenko, Brookes, Berger-Wolf, Stewart, and Stewart]{kabr_telemetry_dataset}
Jenna Kline, Maksim Kholiavchenko, Otto Brookes, Tanya Berger-Wolf, Charles~V. Stewart, and Christopher Stewart.
\newblock Kabr behavior and telemetry dataset, 2024{\natexlab{b}}.
\newblock URL \url{https://imageomics.github.io/KABR/}.

\bibitem[Kline et~al.(2025{\natexlab{a}})Kline, Afridi, Rolland, Maalouf, Laporte-Devylder, Stewart, Crofoot, Stewart, Rubenstein, and Berger-Wolf]{kline2025studying}
Jenna Kline, Saadia Afridi, Edouard~GA Rolland, Guy Maalouf, Lucie Laporte-Devylder, Christopher Stewart, Margaret Crofoot, Charles~V Stewart, Daniel~I Rubenstein, and Tanya Berger-Wolf.
\newblock Studying collective animal behaviour with drones and computer vision.
\newblock \emph{Methods in Ecology and Evolution}, 2025{\natexlab{a}}.
\newblock URL \url{https://doi.org/10.1111/2041-210X.70128}.

\bibitem[Kline et~al.(2025{\natexlab{b}})Kline, Kholiavchenko, Ramirez, Stevens, Babu, Banerji, Campolongo, Thompson, Tiel, Miliko, Duporge, Rosser, Stewart, Berger-Wolf, and Rubenstein]{kline2025kabr-tools-methodology}
Jenna Kline, Maksim Kholiavchenko, Michelle Ramirez, Sam Stevens, Reshma~Ramesh Babu, Namrata Banerji, Elizabeth Campolongo, Matthew Thompson, Nina~Van Tiel, Jackson Miliko, Isla Duporge, Neil Rosser, Chuck Stewart, Tanya Berger-Wolf, and Daniel Rubenstein.
\newblock kabr-tools methodology dataset, 2025{\natexlab{b}}.
\newblock URL \url{https://huggingface.co/datasets/imageomics/kabr-methodology}.

\bibitem[Kline et~al.(2025{\natexlab{c}})Kline, Kholiavchenko, Ramirez, Stevens, Sheets, Babu, Banerji, Campolongo, Thompson, Tiel, Miliko, Rosser, Duporge, Stewart, Berger-Wolf, and Rubenstein]{KABR_Raw_Videos}
Jenna Kline, Maksim Kholiavchenko, Michelle Ramirez, Samuel Stevens, Alec Sheets, Reshma~Ramesh Babu, Namrata Banerji, Elizabeth Campolongo, Matthew Thompson, Nina~Van Tiel, Jackson Miliko, Neil Rosser, Isla Duporge, Charles Stewart, Tanya Berger-Wolf, and Daniel Rubenstein.
\newblock Kabr raw videos: Unprocessed drone footage for kenyan animal behavior analysis (revision fda789c), 2025{\natexlab{c}}.
\newblock URL \url{https://huggingface.co/datasets/imageomics/KABR-raw-videos}.

\bibitem[Kline et~al.(2025{\natexlab{d}})Kline, Zhong, Irizarry, Stewart, Stewart, Rubenstein, and Berger-Wolf]{kline2025wildwing}
Jenna Kline, Alison Zhong, Kevyn Irizarry, Charles~V Stewart, Christopher Stewart, Daniel~I Rubenstein, and Tanya Berger-Wolf.
\newblock Wildwing: An open-source, autonomous and affordable uas for animal behaviour video monitoring.
\newblock \emph{Methods in Ecology and Evolution}, 2025{\natexlab{d}}.
\newblock URL \url{https://doi.org/10.1111/2041-210X.70018}.

\bibitem[Kline et~al.(2025{\natexlab{e}})Kline, Zhong, Kholiavchenko, Stevens, Sheets, Babu, Banerji, Campolongo, Thompson, Van~Tiel, Miliko, Rosser, Stewart, Berger-Wolf, and Rubenstein]{KABR_worked_example}
Jenna Kline, Alison Zhong, Maksim Kholiavchenko, Sam Stevens, Alec Sheets, Reshma Babu, Namrata Banerji, Elizabeth Campolongo, Matthew Thompson, Nina Van~Tiel, Jackson Miliko, Neil Rosser, Charles Stewart, Tanya Berger-Wolf, and Daniel Rubenstein.
\newblock Kabr worked example: Manually annotated detections and behavioral analysis for kenyan wildlife pipeline demonstration, 2025{\natexlab{e}}.
\newblock URL \url{https://huggingface.co/datasets/imageomics/kabr-worked-examples}.

\bibitem[Koger et~al.(2023)Koger, Deshpande, Kerby, Graving, Costelloe, and Couzin]{Koger_Deshpande_Kerby_Graving_Costelloe_Couzin_2023}
Benjamin Koger, Adwait Deshpande, Jeffrey~T. Kerby, Jacob~M. Graving, Blair~R. Costelloe, and Iain~D. Couzin.
\newblock Quantifying the movement, behaviour and environmental context of group-living animals using drones and computer vision.
\newblock \emph{Journal of Animal Ecology}, 92\penalty0 (7):\penalty0 1357–1371, 2023.
\newblock ISSN 1365-2656.
\newblock \doi{10.1111/1365-2656.13904}.

\bibitem[Li et~al.(2022)Li, Wang, He, Li, Wang, Wang, and Qiao]{li2022uniformerv2}
Kunchang Li, Yali Wang, Yinan He, Yizhuo Li, Yi~Wang, Limin Wang, and Yu~Qiao.
\newblock Uniformerv2: Spatiotemporal learning by arming image vits with video uniformer, 2022.

\bibitem[Lin et~al.(2015)Lin, Maire, Belongie, Bourdev, Girshick, Hays, Perona, Ramanan, Zitnick, and Dollár]{coco2015}
Tsung-Yi Lin, Michael Maire, Serge Belongie, Lubomir Bourdev, Ross Girshick, James Hays, Pietro Perona, Deva Ramanan, C.~Lawrence Zitnick, and Piotr Dollár.
\newblock Microsoft coco: Common objects in context, 2015.

\bibitem[Newton-Fisher(2021)]{animalbehaviourpro}
Nicholas~E. Newton-Fisher.
\newblock Animal behaviour pro: 1.6, September 2021.
\newblock URL \url{https://kar.kent.ac.uk/82465/}.

\bibitem[Pedrazzi et~al.(2025)Pedrazzi, Naik, Sandbrook, Lurgi, F{\"u}rtbauer, and King]{pedrazzi2025advancing}
Lucia Pedrazzi, Hemal Naik, Chris Sandbrook, Miguel Lurgi, Ines F{\"u}rtbauer, and Andrew~J King.
\newblock Advancing animal behaviour research using drone technology.
\newblock \emph{Animal Behaviour}, 222:\penalty0 123147, 2025.

\bibitem[Petso et~al.(2021)Petso, Jamisola~Jr, Mpoeleng, Bennitt, and Mmereki]{petso2021automatic}
Tinao Petso, Rodrigo~S Jamisola~Jr, Dimane Mpoeleng, Emily Bennitt, and Wazha Mmereki.
\newblock Automatic animal identification from drone camera based on point pattern analysis of herd behaviour.
\newblock \emph{Ecological informatics}, 66:\penalty0 101485, 2021.

\bibitem[Riginos(2015)]{riginos2015climate}
Corinna Riginos.
\newblock Climate and the landscape of fear in an african savanna.
\newblock \emph{Journal of Animal Ecology}, 84\penalty0 (1):\penalty0 124--133, 2015.

\bibitem[Rolnick et~al.(2024)Rolnick, Aspuru-Guzik, Beery, Dilkina, Donti, Ghassemi, Kerner, Monteleoni, Rolf, Tambe, et~al.]{rolnick2024application}
David Rolnick, Alan Aspuru-Guzik, Sara Beery, Bistra Dilkina, Priya~L Donti, Marzyeh Ghassemi, Hannah Kerner, Claire Monteleoni, Esther Rolf, Milind Tambe, et~al.
\newblock Application-driven innovation in machine learning.
\newblock \emph{arXiv preprint arXiv:2403.17381}, 2024.

\bibitem[Rubenstein et~al.(2016)Rubenstein, Low~Mackey, Davidson, Kebede, and King]{rubenstein2016equus}
D.~Rubenstein, B.~Low~Mackey, ZD~Davidson, F.~Kebede, and S.~R.~B. King.
\newblock Equus grevyi.
\newblock The IUCN Red List of Threatened Species 2016: e.T7950A89624491, 2016.
\newblock https://dx.doi.org/10.2305/IUCN.UK.2016-3.RLTS.T7950A89624491.en, Accessed on 06 August 2025.

\bibitem[Rubenstein(1986)]{rubenstein1986ecology}
D.~I. Rubenstein.
\newblock Ecology and sociality in horses and zebras.
\newblock In D.~I. Rubenstein and R.~W. Wrangham, editors, \emph{Ecological Aspects of Social Evolution}, pages 282--302. Princeton University Press, Princeton, NJ, 1986.

\bibitem[Rubenstein(2010)]{rubenstein2010ecology}
Daniel~I Rubenstein.
\newblock Ecology, social behavior, and conservation in zebras.
\newblock In \emph{Advances in the Study of Behavior}, volume~42, pages 231--258. Elsevier, 2010.

\bibitem[Saoud et~al.(2024)Saoud, Sultan, Elmezain, Heshmat, Seneviratne, and Hussain]{saoud2024beyond}
Lyes~Saad Saoud, Atif Sultan, Mahmoud Elmezain, Mohamed Heshmat, Lakmal Seneviratne, and Irfan Hussain.
\newblock Beyond observation: Deep learning for animal behavior and ecological conservation.
\newblock \emph{Ecological Informatics}, page 102893, 2024.

\bibitem[Schad and Fischer(2023)]{schad2023opportunities}
Lukas Schad and Julia Fischer.
\newblock Opportunities and risks in the use of drones for studying animal behaviour.
\newblock \emph{Methods in Ecology and Evolution}, 14\penalty0 (8):\penalty0 1864--1872, 2023.

\bibitem[Smith and Pinter-Wollman(2021)]{smith2021observing}
Jennifer~E Smith and Noa Pinter-Wollman.
\newblock Observing the unwatchable: Integrating automated sensing, naturalistic observations and animal social network analysis in the age of big data.
\newblock \emph{Journal of Animal Ecology}, 90\penalty0 (1):\penalty0 62--75, 2021.

\bibitem[Zhang et~al.(2024)Zhang, Zhao, Fu, Luo, Shao, Zhang, and Yu]{zhang2024reliable}
Guoqing Zhang, Yongxiang Zhao, Ping Fu, Wei Luo, Quanqin Shao, Tongzuo Zhang, and Zhongde Yu.
\newblock A reliable unmanned aerial vehicle multi-target tracking system with global motion compensation for monitoring procapra przewalskii.
\newblock \emph{Ecological Informatics}, 81:\penalty0 102556, 2024.

\end{thebibliography}

\newpage

\section{Appendix}
\label{appendix}


\subsection{Field protocols}
\label{app:fieldprotocol}
We manually recorded ground-based behavioral data of zebras and giraffes at the Mpala Research Center to assess the consistency between drone-based and traditional behavioral sampling methods. We visually located the herds on the ground from the field vehicle, launched the drone, and began manual ground-based behavioral observations once the animals were visible from both the drone and the field vehicle.

\subsubsection{Ground-based observation protocols}
Field observations were conducted by trained observers who underwent rigorous calibration to ensure data quality and consistency. Observers participated in field calibration exercises, practicing simultaneous observations overseen by an expert field ecologist. Throughout data collection, teams conducted daily 15-minute synchronization sessions to maintain consistency and address any emerging discrepancies in behavioral classification. Each observation team utilized 10 X 42 binoculars for detailed behavioral observation. We used the iOS version of the Animal Behaviour Pro \citep{animalbehaviourpro} to record timestamped behavior observations data synchronized with GPS coordinates on our iPhone mobile devices.

We used two complementary sampling methods to capture different aspects of behavioral data: scan sampling and focal sampling. Scan sampling sessions targeted groups of 3-12 individuals observed for an average duration of 10.2 $\pm$ 2.3 minutes. During these sessions, instantaneous scans were conducted every 2 minutes, with observers systematically recording the behavior of all visible individuals from left to right across the group. This approach provided population-level behavioral data while controlling for environmental and social context. Focal sampling was conducted simultaneously by a second observer team, with continuous observation of single individuals for 10-minute periods. Focal animals were selected based on clear visibility and distinctive identifying features such as unique stripe patterns for zebras or spot configurations for giraffes. All behavioral transitions were recorded with 1-second precision, providing fine-scale temporal resolution of behavioral sequences. We completed six simultaneous scan and focal sessions across different groups and species, providing paired datasets for methodological comparison.

\subsubsection{Drone protocol}

To minimize behavioral disturbance, we employed standardized approach protocols. We visually located the herds and slowly approached in out field vehicle until we were positioned approximately 100 meters away with the takeoff location positioned with the vehicle between the pilot and the animals. After observing the animals for five to ten minutes to gauge vigilance, the drone pilot exited the vehicle on the opposite side from the animals to position the drone for takeoff.
After takeoff, the drone's altitude was set immediately to 40-50 meters. The drone hovered in place for 5 minutes to gauge whether the animals reacted. If the animals appeared undisturbed, the pilot gradually decreased altitude to 20-30 meters. Note, this is lower than typical wildlife drone surveys (70-100m) to maintain sufficient video resolution for individual behavioral identification, and hovered briefly before slowly approaching the group. We immediately paused drone operations whenever animals exhibited signs of agitation or increased vigilance behavior. This conservative approach prioritized animal welfare while maintaining data quality for behavioral analysis. The drone's camera was angled approximately 45 degrees with respect to the terrain to better capture the side-view of the animals, with flight speeds kept below 5 m/s when tracking moving groups to maintain smooth footage. Our battery management protocols ensured maximum 20-minute flights with 20\% power reserve for safe return. All videos included embedded GPS coordinates and timestamps enabling precise synchronization with ground observations. We uploaded full-length drone videos and corresponding mission telemetry data on the Ohio Supercomputer Center (OSC) for storage and processing.

\subsection{Demographic Information}

\begin{table*}[h]
    \centering
    \begin{tabular}{llcc}
        \toprule
        \textbf{Session} & \textbf{Species} & \textbf{Male} & \textbf{Female}  \\
        \midrule
        1 & Plains Zebra & 1 & 2  \\
        2 & Plains Zebra & 1 & 4  \\
        3 & Grevy's Zebra & 3 & 0 \\
        4 & Grevy's Zebra & 3 & 0 \\
        5 & Reticulated giraffe & 1 & 1 \\
        6 & Reticulated giraffe & 2 & 0\\
        \bottomrule
    \end{tabular}
    \caption{Sessions capturing behavior with simultaneous scan and focal sampling, \citep{kline2025kabr-tools-methodology}. Two simultaneous scan and focal sampling sessions were captured for each of the three ungulate species with a mix of male and females in each group. Simultaneous scan and focal sampling were collected to compare the relative strengths and weaknesses of each methodology.}
    \label{tab:scan_focal}
\end{table*}

\begin{table*}[h]
\centering
\begin{tabular}
{p{1cm}p{11cm}p{2cm}}
    \toprule
    \centering
    \textbf{Zebra ID} & \textbf{Demographic information}  & \textbf{Temporal Overlap}  \\
    \midrule
    A& Adult female lactating Grevy’s zebra travelling with her foal and two female plains zebras &  96 sec\\
    B& Adult female Grevy’s zebra, travelling with Grevy’s herd &  141 sec\\
    C& Female foal Grevy’s zebra, travelling with Grevy’s herd &  197 sec\\
    D&Plains male zebra, travelling with plains harem & 72 sec\\
 \bottomrule
\end{tabular}
\caption{Demographic details of individual zebras' behaviors analyzed to compare ground focal with drone focal annotations, \cite{kline2025kabr-tools-methodology}. The temporal overlap quantifies the time when ground and drone focal sampling coincide.}
\label{table:focal_ind_zebras}
\end{table*}

\begin{table*}[h]
\centering
\renewcommand{\arraystretch}{0.85}
\begin{tabular}{cccc}
\toprule
\textbf{Session} & \textbf{Habitat} & \textbf{Herd Size Category (Count)} & \textbf{Date} \\
\midrule
1 & Closed & Small (3) & 11/01/23 \\
2 & Open & Small (3) &  11/01/23 \\
3 & Open --> Closed & Small (3) &  12/01/23 \\
4 & Closed & Large (7) & 16/01/23 \\
5 & Open & Small (2) & 18/01/23 \\
6 & Open & Large (5) & 20/01/23 \\
\bottomrule \\
\end{tabular}
\caption{Summary of data used for Case Study 1: Grevy's Zebras Landscape of Fear, by habitat and herd size, \citep{kline_kabr_telemetry, KABR_worked_example}. Herds larger than the median (4.2) are considered `Large', all others are considered `Small'. For session 3, the habitat type transitioned from Open to Closed as the animals moved to a more dense habitat during the video duration. }
\label{tab:habitat_data}
\end{table*}

\newpage
\subsection{Machine learning performance}
\label{app_ml}

\begin{table*}[h]
\centering
\footnotesize
\begin{tabular*}{\textwidth}{@{\extracolsep{\fill}}lcccccc@{}}
\toprule
\multirow{2}{*}{\textbf{Model}} &
\multicolumn{3}{c}{\textbf{Mean Average Precision (mAP) (\%)}} & \multirow{2}{*}{\textbf{Precision}} & \multirow{2}{*}{\textbf{Recall}} & \multirow{2}{*}{\textbf{F1}} \\
\cmidrule{2-4}
& Overall & Head Classes & Tail Classes & & & \\
\midrule
I3D & 65.06 & 96.81 & 54.48 & \textbf{67.17} & 62.94 & 64.52 \\
SlowFast & 66.10 & 96.72 & 55.90 & 67.05 & \textbf{65.28} & \textbf{65.82} \\
X3D-L & \textbf{66.36} & \textbf{96.96} & \textbf{56.16} & 66.44 & 63.65 & 64.70 \\
UniformerV2-B & 61.78 & 95.38 & 50.58 & 64.37 & 54.82 & 57.41 \\
\bottomrule
\end{tabular*}
\caption{Behavior recognition machine learning model performance, summarized from \citep{kabrdeepdive}. The mean average precision (mAP \%) is reported overall (all 8 classes), Head classes (\textit{Walk}, \textit{Graze}), and Tail classes (\textit{Head Up, Browse, Auto-Groom, Trot, Run, Out of Sight/Occluded}). The precision (P), recall (R), and F1 score (F1) are reported for each model. The highest scores across all metrics are shown in \textbf{bold} for each evaluation setting. Refer to our previously published work for additional model training and performance details  \citep{kabrdeepdive}.}
\label{table:mlmodel}
\end{table*}

\begin{table}[htbp]
\centering
\begin{tabular}{ll}
\toprule
\textbf{Metric} & \textbf{Formula} \\
\midrule
Precision & $\displaystyle \frac{\text{TP}}{\text{TP} + \text{FP}}$ \\
Recall & $\displaystyle \frac{\text{TP}}{\text{TP} + \text{FN}}$ \\
F1 Score & $\displaystyle 2 \times \frac{\text{Precision} \times \text{Recall}}{\text{Precision} + \text{Recall}}$ \\
Average Precision (AP) & Area under the Precision-Recall curve \\
Mean Average Precision (mAP) & $\displaystyle \frac{1}{N} \sum_{i=1}^{N} \text{AP}_i$ \\
\bottomrule  \\
\end{tabular}
\caption{Summary of evaluation metrics used in object detection and classification. TP = True Positives, FP = False Positives, FN = False Negatives, $N$ = Number of classes, and $\text{AP}_i$ = Average Precision for class $i$.}
\label{tab:evaluation-metrics}
\end{table}

\newpage
\subsection{Ethogram}
\label{ethogram}
\begin{table}[h]
\centering
\footnotesize
\begin{tabular}{p{2.5cm}p{0.8cm}p{0.8cm}p{6cm}}
\toprule
\textbf{Behaviour} & \textbf{Code} & \textbf{Species} & \textbf{Notes} \\
\midrule
Auto-groom & AG/SG & Both & Self grooming, rubbing nose or teeth on the neck, back or other body parts, or rubbing body on a landscape feature, such as a tree or branch \\
Browsing & B & G & Eating leaves off trees, chewing and looking around \\
Chasing & C & Z & Focal animal chasing another animal of same species \\
Defecating & DF/D & Both & Elimination of solid waste\\
Drinking & D & Z & Ingesting water \\
Dusting & DU & Z & Rolling on back to covered body in dust \\
Fighting & F & Both & Aggression towards another animal, involves biting, rearing and kicking with fore or hind legs \\
Graze & G & Z &  Head down, biting and chewing grass\\
Head Up & HU & Both & Vigilance, standing, sleeping, or scanning \\
Herding & H & Z & Head down with neck stretched out, directing another animals' movements\\
Lying down & L & Z & Body lying on ground\\
Mounting/Mating & M & Both & Engaging in mating behaviors, such as mounting \\
Mutual grooming & MG & Both & Rubbing with nose or teeth each others neck, back or other body parts\\
Running & R & Both & A high-speed gait where both diagonal pairs of legs move together, but there is a moment when all four hooves are off the ground.\\
Sniff & S & Z & Sniffing another animal or nuzzling the ground \\
Trotting & TR & Z &  A faster, more energetic gait where diagonal pairs of legs (front left with back right, and vice versa) move together, resulting in a brief period when all four hooves are off the ground. \\
Urinating & U & Both & Elimination of liquid waste \\
Walking & W & Both & A slow, steady gait where one leg moves at a time, with at least one hoof in contact with the ground at all times \\
\midrule
\multicolumn{4}{l}{\textit{Technical annotations}} \\
Occluded$^*$ & OCL & Both & Animal hidden behind bush/another animal \\
Out of Focus$^*$ & OOC & Both & Object detector switches to wrong animal \\
Out of Frame$^*$ & OOF & Both & Animal not visible in video frame \\
\bottomrule \\
\end{tabular}
\caption{Combined ethogram for zebras and giraffes. Species column indicates applicability: Z = zebras only, G = giraffes only, Both = applicable to both species. $^*$Technical categories can overlap (e.g., detector error may precede animal going out of frame).}
\label{tab:combined_ethogram}
\end{table}





\end{document}